\documentclass[11pt]{article}

\usepackage[final]{acl}

\usepackage{times}
\usepackage{latexsym}

\usepackage[T1]{fontenc}

\usepackage[utf8]{inputenc}

\usepackage{microtype}

\usepackage{inconsolata}

\usepackage{graphicx}
\usepackage{booktabs}

\usepackage{multirow}
\usepackage{array}
\usepackage{xcolor}
\usepackage{colortbl}
\usepackage{hyperref}
\usepackage{multirow}
\title{MegaRAG: Multimodal Knowledge Graph-Based \\ Retrieval Augmented Generation}

\author{
\textbf{Chi-Hsiang Hsiao}\textsuperscript{1,}\thanks{Equal contribution.} \quad
\textbf{Yi-Cheng Wang}\textsuperscript{1,}\footnotemark[1] \quad
\textbf{Tzung-Sheng Lin}\textsuperscript{4} \\
\textbf{Yi-Ren Yeh}\textsuperscript{2} \quad
\textbf{Chu-Song Chen}\textsuperscript{1,3} \\
\textsuperscript{1}Department of Computer Science and Information Engineering, National Taiwan University \\
\textsuperscript{2}Department of Mathematics, National Kaohsiung Normal University \\
\textsuperscript{3}FinTech Center, National Taiwan University \\
\textsuperscript{4}E.SUN Financial Holding Co., Ltd. \\
\textsuperscript{1}\texttt{\{r12922048, d13922033, chusong\}@csie.ntu.edu.tw} \\
\textsuperscript{4}\texttt{francis-17710@esunbank.com} \quad
\textsuperscript{2}\texttt{yryeh@nknu.edu.tw} \\
}

\begin{document}
\maketitle
\begin{abstract}
Retrieval-augmented generation (RAG) enables large language models (LLMs) to dynamically access external information, which is powerful for answering questions over previously unseen documents. Nonetheless, they struggle with high-level conceptual understanding and holistic comprehension due to limited context windows, which constrain their ability to perform deep reasoning over long-form, domain-specific content such as full-length books. To solve this problem, knowledge graphs (KGs) have been leveraged to provide entity-centric structure and hierarchical summaries, offering more structured support for reasoning. However, existing KG-based RAG solutions remain restricted to text-only inputs and fail to leverage the complementary insights provided by other modalities such as vision. On the other hand, reasoning from visual documents requires textual, visual, and spatial cues into structured, hierarchical concepts. To address this issue, we introduce a multimodal knowledge graph-based RAG that enables cross-modal reasoning for better content understanding. Our method incorporates visual cues into the construction of knowledge graphs, the retrieval phase, and the answer generation process. Experimental results across both global and fine-grained question answering tasks show that our approach consistently outperforms existing approaches on both textual and multimodal benchmarks. Our code is available on \href{https://github.com/AI-Application-and-Integration-Lab/MegaRAG}{\textit{https://github.com/AI-Application-and-Integration-Lab/MegaRAG.}}
\end{abstract}

\section{Introduction}
Humans naturally integrate multiple modalities such as textual, visual, and layout to fluidly transition between abstract and detailed reasoning. However, multimodal large language models (MLLMs) \cite{bai2025qwen2,grattafiori2024llama,hurst2024gpt,team2023gemini}, despite recent progress, remain limited by constrained context windows, restricting their ability to process long-form, domain-specific content. Interpreting a history textbook involves both global and localized observations, which remains challenging for MLLMs.

On the other hand, RAG can enhance LLMs by providing on-demand access to external knowledge. Early text-based RAG relied on sparse or dense retrieval but struggled with deep, multi-hop reasoning in multimodal documents. Recently, Graph-based RAG introduces structured abstraction via entity-relation graphs. With models like GraphRAG \cite{edge2024local} and LightRAG \cite{guo2024lightrag}, long-range knowledge retrieval of improved scalability are enhanced through KG-assisted retrieval pipelines. However, these methods excel in text-based multi-hop reasoning but remain constrained in handling complex, multimodal content.
Current graph-based RAG methods face some key limitations. First, existing approaches remain unimodal, overlooking visual cues like diagrams, charts or maps, yielding disjointed representations that hinder multimodal reasoning. Additionally, due to context window constraints, most approaches segment documents into independent chunks, extracting entities separately rather than sequentially. This leads to fragmented KGs that miss cross-chunk relationships and key entities. 

To our knowledge, while recent studies have explored manually constructed multimodal knowledge graphs (KGs) for RAG-based question answering \cite{lee-etal-2024-multimodal}, automatically building such KGs for RAG-assisted reasoning remains underexplored. To address this gap, we introduce MegaRAG, a multimodal, graph-based RAG method that enhances cross-modal reasoning.

To better handle the association of different modalities in visual documents, more relations beyond text-to-texts need to be extracted, such as text-to-figures and figure-to-figure relations. Although the parallel-reading-then-combining strategy can refine entities and relations as in GraphRAG~\cite{edge2024local} and LightRAG~\cite{guo2024lightrag}, such refinement still relies on a single chunk while overlooking global document information. To address this limitation, we design a page-based, two-round approach for KG construction. Our solution initiates a KG by simply extracting entity-relation pairs in parallel for every page of a document using existing MLLMs, and the page-based relations are joined to form an initial graph. As the initial KG may not capture the inter-relationship between texts and visual elements sufficiently well, we conduct refinement processes in subsequent stage(s), where the initial KG(s) serve as global guidance to capture subtle relationships often lost in naïve, isolated extraction. In particular, to maintain scalability while incorporating long-range dependencies, we avoid injecting the entire initial KG into the MLLM inputs. Instead, we retrieve only a subgraph of the entire KG for each page, yielding a lightweight yet context-aware input. This strategy enables progressive improvement of the graph’s structural coherence and cross-modal grounding. 

We validate MegaRAG across global (book-level) and local (page/slide-level) QA benchmarks, spanning both text-only and multimodal datasets. Experimental results demonstrate that MegaRAG consistently outperforms strong baselines, particularly in scenarios requiring deep cross-modal integration and structured abstraction. Our contributions are summarized as follows.

\noindent $\bullet$~~We introduce MegaRAG, an easy-to-use system that automatically constructs Multimodal KGs for visual document question answering with MLLMs.\\
\noindent $\bullet$~~We develop a novel refinement process that enhances cross-modal grounding while addressing limitations in independent KG construction.\\
\noindent $\bullet$~~We demonstrate that MegaRAG outperforms strong baselines on both global and local QA tasks, including GraphRAG and LightRAG.

\begin{figure*}[!t]
    \centering
    \includegraphics[width=0.85\linewidth]{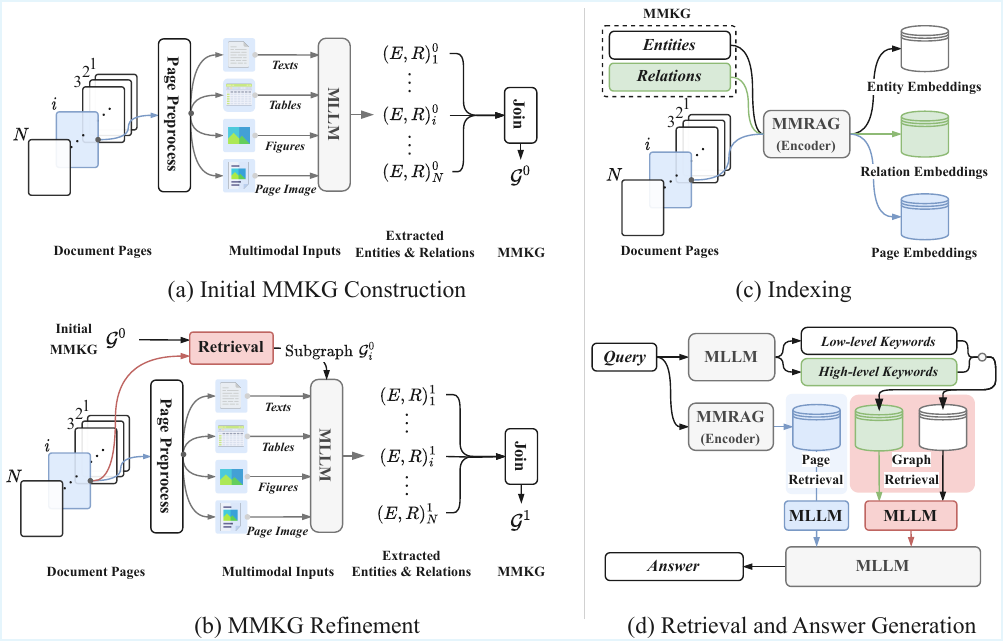}
    \caption{
        Overview of our MegaRAG for MMKG construction and MMKG-augmented generation.
        (a) Initial MMKG construction: an MLLM extracts page-level entities/relations $(E,R)^0_i$ in parallel and merges them into $\mathcal{G}^0$.
        (b) Refinement: each page retrieves a subgraph $\mathcal{G}^0_i$ to refine the graph, producing $\mathcal{G}^1$.
        (c) Indexing: $\mathcal{G}^1$ and page images are encoded into dense entity, relation, and page embeddings.
        (d) Retrieval \& generation: low-/high-level keywords retrieve relevant subgraphs/pages, followed by two-stage answer generation.
    }
    \label{fig:main_diagram}
\end{figure*}

\section{Related Work}
\label{related_work}
We briefly review several major directions of RAG: including retrieving information directly from raw data sources such as documents and images, and integrating structured knowledge through KGs.

\noindent\textbf{RAG with Raw Data Source.}
Early RAG methods \cite{guu2020retrieval,lewis2020retrieval} retrieve text chunks from corpora to support answer generation, primarily relying on retrieval strategies either sparse or dense. Sparse methods exemplified by TF-IDF \cite{10.1145/361219.361220} and BM25 \cite{10.1561/1500000019} depend on lexical heuristics to match queries with relevant text segments. They offer computational efficiency but lack deeper semantic comprehension. Dense techniques \cite{karpukhin2020dense,khattab2020colbert,santhanam2022colbertv2} project queries and documents into a shared embedding space, significantly improving retrieval performance of lexical variations. 
Subsequent works have enhanced this pipeline using LLM recently: HyDE \cite{gao-etal-2023-precise} generates a hypothetical answer to enrich the retrieval query, Self-RAG \cite{asai2024selfrag} introduces reflection tokens to enable adaptive retrieval and self-critique within a single LLM, while RQ-RAG \cite{chan2024rqrag} decomposes the query into sub-queries to improve context coverage.
Despite their strong performance on text-based RAG tasks, these methods often struggle with multimodal documents involving complex texts, layouts and visual elements.

\noindent\textbf{Multimodal RAG (MMRAG).}
To tackle the limitations, more recent studies have focused on multimodal retrieval methods that better retain the structural information of documents. DSE \cite{ma2024unifying} treats document screenshots as unified inputs and directly encodes their visual layout, text, and images into a single vector embedding. ColPaLi~\cite{faysse2024colpali} continues this direction by encoding document images into multi-vector embeddings, effectively capturing fine-grained visual cues. Its variant, ColQwen, replaces the PaLI-Gemma~\cite{beyer2024paligemma} with Qwen2-VL~\cite{wang2024qwen2} and achieves improved retrieval performance. Moving beyond retrieval, VisRAG~\cite{yu2024visrag} integrates MLLMs into the full RAG pipeline. Instead of extracting text, it embeds document images directly for retrieval and incorporates them into the generation stage, allowing the model to jointly reason over visual and textual content.

The above methods excel in text-to-image retrieval but fail to solve tasks involving a mixture of single-modality (e.g., text-to-text), cross-modality (e.g., text-to-image), and fused-modality (text+image-to-text+image) retrieval. GME~\cite{zhang2024gme} tackles this by introducing a unified embedding model that encodes diverse modality combinations and enables flexible retrieval within a shared representation space.

While these approaches significantly enhance document understanding, they neglect the long-range corpus-level structure, which is essential for handling complex, multi-hop QA~\cite{tanaka2023slidevqa,yang2018hotpotqa}. 

\noindent\textbf{RAG with Knowledge Graph.}
Knowledge-augmented generation~\cite{procko2024graph} leverages KGs to provide structured, factual context for LLMs. Within this line of research, SubgraphRAG~\cite{li2025simple} enhances efficiency through lightweight scoring mechanisms for subgraph retrieval, while G-Retriever~\cite{he2024g} frames subgraph selection as a Steiner Tree optimization problem to support large-scale textual graphs. Gao et al.~\cite{gao-etal-2022-graph} employ a learning-to-rank approach to improve retrieval from KGs. While these methods advance graph-based retrieval, they depend on manually constructed KGs, which are costly to build and require substantial domain expertise. Moreover, static KGs are inherently limited in addressing queries that require corpus-level reasoning beyond fixed KGs.

To address this limitation, GraphRAG~\cite{edge2024local} proposes building KGs directly from raw text using LLMs, followed by a hierarchical community detection algorithm~\cite{traag2019louvain} to cluster semantically related nodes. During inference, it prompts the LLM to generate intermediate answers for each community summary, scores them by confidence, and aggregates the top responses into a final answer. Although this enables corpus-level reasoning, it incurs high computational cost due to repeated LLM queries over many community summaries. To improve efficiency, LightRAG~\cite{guo2024lightrag} introduces a two-stage retrieval process: it first extracts local and global keywords from the query, then retrieves relevant nodes and their surrounding subgraphs using dense retrieval. This design reduces the need for repeated LLM inference and significantly improves scalability. which introduces a hybrid RAG framework that alternates between naive and graph-based retrieval. TOG-2~\cite{ma2025thinkongraph} introduces a hybrid RAG method that alternates between dense retrieval and graph reasoning. However, these approaches rely on manually curated KGs, which are costly to construct and limited in coverage.

However, these KG-augmented RAGs rely solely on textual KGs, limiting their ability to handle multimodal content such as images. To overcome this limitation, multimodal knowledge graphs (MMKGs)~\cite{10.1007/978-3-030-21348-0_30,zhangmultimodal} enrich KGs by associating entities with aligned visual (e.g., images), numeric (e.g., dates, measurements), and textual descriptions. A representative benchmark~\cite{10.1007/978-3-030-21348-0_30} introduces MMKGs that were constructed by linking overlapping entities via \textit{sameAS} relations and annotating them with web-crawled images and numeric literals. MMKGs have demonstrated utility across tasks, including KG completion~\cite{mousselly2018multimodal,xie2017image}, recommendation systems~\cite{sun2020multi}, and image captioning~\cite{zhao2023boosting}.

More recently, MMKGs have been integrated into RAG pipelines to support multimodal QA with LLMs. For instance, MR-MKG~\cite{lee-etal-2024-multimodal} utilizes manually constructed MMKGs that encode visual and factual knowledge, enabling LLMs to reason over structured multimodal inputs. Although effective, this approach depends on manually built, domain-specific MMKGs that are costly to scale. Query-driven MMKG~\cite{bu2025query} takes a step toward automatic MMKG construction by dynamically building query-specific local graphs for online multimodal reasoning. Nevertheless, such local graph construction is mainly designed for short-form, query-centered settings and does not explicitly model document-level structure or cross-page dependencies in long-form documents. Therefore, building scalable MMKGs that can be constructed automatically and support open-domain multimodal RAG over long, visually rich documents remains a key challenge.

\section{Methodology}
\label{methodology}
In this section, we present MegaRAG, covering the iterative construction process of MMKG, graph indexing and retrieval mechanisms, and the answer generation pipeline.

\subsection{MMKG Construction}
We define our MMKG as \(\mathcal{G} = (\mathcal{V}, \mathcal{E})\), where \(\mathcal{V}\) is the set of nodes representing entities, and \(\mathcal{E}\) is the set of edges denoting relations between entities. Given a document consisting of \(N\) pages, we extract four types of content from each page \(i\): text content \(\mathrm{T}_i\), figure images \(\mathrm{F}_i\), table images \(\mathrm{B}_i\), and the full-page rendered image \(\mathrm{I}_i\) (which captures the layout of the page). These elements are obtained using a document analysis tool. We define the input for page \(i\) as \(\mathrm{P}_{i} = \{\mathrm{T}_i, \mathrm{F}_i, \mathrm{B}_i, \mathrm{I}_i\}\), which serves as input to our graph construction pipeline.

\noindent\textbf{Initial Graph Construction.}
As illustrated in Figure~\ref{fig:main_diagram}(a), the initial stage involves extracting entities and relations from each page in parallel using a graph generation function \(G(\cdot)\), which leverages an MLLM guided by a task-specific prompt. The prompt specifies the extraction goals, provides reasoning instructions, and enforces a constrained output format to ensure consistency across pages.
In our implementation, GPT-4o-mini serves as the MLLM for the MMKG construction.

Given a multimodal input $\mathrm{P}_i$, the graph generation function produces a set of page-level entities and relations $(\mathrm{E}, \mathrm{R})^{0}_{i} = G(\mathrm{P}_i)$, extracted from both textual and visual content. The MLLM is guided to identify multiple entities within the text and to treat each figure or table as a single, standalone entity. For instance, a bar chart titled ``Monthly Website Visitors'' may be recognized as an entity and connected to surrounding text discussing user engagement trends. Decorative or non-informative visuals, such as background patterns or logos, are ignored. The full-page image $\mathrm{I}_i$ is used solely to support spatial reasoning and does not generate entity nodes. Each extracted entity includes a name, a predefined type (e.g., person, organization), and a description. Relations are defined by a source and target entity, a description, and a set of representative keywords.

After generating the set of page-level entities and relations (denoted as $\{(\mathrm{E}, \mathrm{R})^{0}_{i}\}_{i=1}^{N}$), we merge them into a unified MMKG $\mathcal{G}^{0}$. This involves consolidating entity nodes with the same name and merging relation edges with matching source, target, and relation types. During this process, different descriptions associated with the same entity or relation are aggregated to form a richer, more comprehensive representation. Similarly, keywords from multiple occurrences are accumulated.

\noindent\textbf{Graph Refinement and Enrichment.} 
The initial MMKG \(\mathcal{G}^{0}\) is often incomplete, as many cross-modal entities and relationships may be overlooked during the first-pass extraction. To bridge the gaps, we introduce a refinement stage that enhances graph \(\mathcal{G}^{1}\), leveraging both the original multimodal inputs and the preliminary knowledge encoded in \(\mathcal{G}^{0}\). The process is illustrated in Figure~\ref{fig:main_diagram}(b).

To efficiently refine MMKG under the MLLM's limited context window, we focus on constructing lightweight, page-specific subgraphs rather than processing the entire graph. For each page \(i\), we extract a context-specific subgraph \(\mathcal{G}^{0}_{i}\) from \(\mathcal{G}^{0}\). In practice, we reuse entity names and relation keywords from the previously extracted page-level output \({(\mathrm{E}, \mathrm{R})}^{0}_{i}\) to retrieve relevant content in \(\mathcal{G}^{0}\), reducing redundancy and simplifying subgraph construction. These entity names and relation keywords are encoded into semantic embeddings and efficiently matched against dense vector representations of entities and relations built from initial MMKG. To enrich the local context, the selected nodes and edges are further expanded by including their one-hop neighbors, resulting in a compact yet informative subgraph. 
A detailed explanation of this graph indexing and retrieval process is provided in Section~\ref{sec:indexing_retrival}.

The refinement process is formalized as \((\mathrm{E}, \mathrm{R})^{1}_{i} = R(\mathrm{P}_{i}, \mathcal{G}^{0}_{i})\), where \(R(\cdot)\) is a refinement function that reuses the same MLLM from the initial stage, now guided by a KG-specific refinement prompt. 
Since the pages remain independent when extracting the entity relationship leveraging the subgraph, the benefit of parallelism is maintained for efficient graph construction.
This function identifies missing knowledge in page \( \mathrm{P}_{i} \) by examining the retrieved subgraph \( \mathcal{G}^{0}_{i} \). Specifically, it detects entities mentioned in the input that are not yet present in the subgraph, as well as implicit relations between entities that are suggested by the content but missing from \( \mathcal{G}^{0}_{i} \).

For example, consider a page where the text states ``Electric vehicle sales increased significantly in 2023,'' and a nearby figure titled ``Annual Sales by Vehicle Type'' presents a bar chart with a prominent ``EV'' bar (denoting Electric Vehicles). In the initial extraction, the text and the figure may be treated as independent entities. During refinement, the MLLM infers that the figure visually supports the textual claim and adds a relation such as \textit{illustrates} or \textit{supports} between the textual entity ``Electric vehicle sales in 2023'' and the visual entity ``Annual Sales by Vehicle Type.'' 

These newly identified entities and relations are added to the refined set \((\mathrm{E}, \mathrm{R})^{1}_{i}\). The updated page-level outputs \(\{(\mathrm{E}, \mathrm{R})^{1}_{i}\}_{i=1}^{N}\) are then merged to form the enriched MMKG \(\mathcal{G}^{1}\). Although we perform only a single refinement step, the process can be applied iteratively to further improve graph completeness. To balance effectiveness and efficiency, we adopt one round of refinement and provide the full prompt formats used for both the initial construction and refinement.
More details can be found in Appendix~\ref{sec:appendix_implementation_details}.

\begingroup
\setlength{\tabcolsep}{8pt}     
\renewcommand{\arraystretch}{1.1} 

\begin{table*}
\centering
\resizebox{\linewidth}{!}{%
\begin{tabular}{lcccccccccccc} 
\toprule
 & \multicolumn{3}{c}{\textbf{Agriculture}} & \multicolumn{3}{c}{\textbf{CS}} & \multicolumn{3}{c}{\textbf{Legal}} & \multicolumn{3}{c}{\textbf{Mix}} \\ 
\cmidrule(r){2-4}\cmidrule(lr){5-7}\cmidrule(r){8-10}\cmidrule(lr){11-13}
 & NaiveRAG & \textbf{Ours} & Tie & NaiveRAG & \textbf{\textbf{Ours}} & Tie & NaiveRAG & \textbf{\textbf{Ours}} & Tie & NaiveRAG & \textbf{\textbf{Ours}} & Tie \\ 
\hline
Comprehensiveness & 5.6 & \textbf{42.7} & 51.6 & 7.2 & \textbf{44.8} & 48.0 & 8.0 & \textbf{51.2} & 40.8 & 5.6 & \textbf{50.4} & 44.0 \\
Diversity & 16.1 & \textbf{70.2} & 13.7 & 14.4 & \textbf{68.0} & 17.6 & 17.6 & \textbf{69.6} & 12.8 & 12.0 & \textbf{77.6} & 10.4 \\
Empowerment & 12.9 & \textbf{66.9} & 20.2 & 22.4 & \textbf{47.2} & 30.4 & 20.8 & \textbf{61.6} & 17.6 & 20.0 & \textbf{62.4} & 17.6 \\
Overall & 8.1 & \textbf{62.1} & 29.8 & 9.6 & \textbf{53.6} & 36.8 & 13.6 & \textbf{64.0} & 22.4 & 8.0 & \textbf{66.4} & 25.6 \\ 
\cmidrule(lr){2-4}\cmidrule(lr){5-7}\cmidrule(lr){8-10}\cmidrule(lr){11-13}
 & GraphRAG & \textbf{\textbf{Ours}} & Tie & GraphRAG & \textbf{\textbf{Ours}} & Tie & GraphRAG & \textbf{\textbf{Ours}} & Tie & GraphRAG & \textbf{\textbf{Ours}} & Tie \\ 
\hline
Comprehensiveness & 5.6 & \textbf{64.8} & 29.6 & 4.0 & \textbf{68.0} & 28.0 & 11.2 & \textbf{60.8} & 28.0 & 4.0 & \textbf{59.2} & 36.8 \\
Diversity & 14.4 & \textbf{76.8} & 8.8 & 14.4 & \textbf{72.0} & 13.6 & 12.8 & \textbf{75.2} & 12.0 & 22.4 & \textbf{59.2} & 18.4 \\
Empowerment & 0.8 & \textbf{94.4} & 4.8 & 2.4 & \textbf{93.6} & 4.0 & 10.4 & \textbf{86.4} & 3.2 & 12.0 & \textbf{80.0} & 8.0 \\
Overall & 5.6 & \textbf{82.4} & 12.0 & 4.0 & \textbf{79.2} & 16.8 & 11.2 & \textbf{80.0} & 8.8 & 7.2 & \textbf{70.4} & 22.4 \\ 
\cmidrule(lr){2-4}\cmidrule(lr){5-7}\cmidrule(lr){8-10}\cmidrule(lr){11-13}
 & LightRAG & \textbf{\textbf{Ours}} & Tie & LightRAG & \textbf{\textbf{Ours}} & Tie & LightRAG & \textbf{\textbf{Ours}} & Tie & LightRAG & \textbf{\textbf{Ours}} & Tie \\ 
\hline
Comprehensiveness & 4.0 & \textbf{65.6} & 30.4 & 3.2 & \textbf{68.8} & 28.0 & 9.6 & \textbf{54.4} & 36.0 & 3.2 & \textbf{76.8} & 20.0 \\
Diversity & 10.4 & \textbf{70.4} & 19.2 & 12.8 & \textbf{72.0} & 15.2 & 14.4 & \textbf{69.6} & 16.0 & 11.2 & \textbf{76.8} & 12.0 \\
Empowerment & 4.8 & \textbf{76.0} & 19.2 & 10.4 & \textbf{75.2} & 14.4 & 12.0 & \textbf{73.6} & 14.4 & 4.0 & \textbf{80.8} & 15.2 \\
Overall & 4.8 & \textbf{75.2} & 20.0 & 4.8 & \textbf{76.8} & 18.4 & 11.2 & \textbf{72.0} & 16.8 & 7.2 & \textbf{80.0} & 12.8 \\
\bottomrule
\end{tabular}
}
\caption{Performance on the UltraDomain benchmark in terms of win rates (\%).}
\label{tab:main_result_ultradomain}
\end{table*}

\endgroup

\subsection{Indexing and Retrieval}
\label{sec:indexing_retrival}

We adopt a unified retrieval framework that integrates graph structure, represented by entities and relations, along with page images within a shared embedding space to enable seamless cross-modal retrieval. Specifically, we use GME~\cite{zhang2024gme}, a multimodal encoder that jointly embeds textual and visual inputs. GME aligns all content types, including both textual and visual information, into a common vector space, supporting text-to-text and text-to-image retrieval through a unified representation.

\noindent\textbf{Indexing.}
Our indexing process encompasses three content types, as illustrated in Figure~\ref{fig:main_diagram}(c): document page images, entities, and relations. Page images are directly encoded using GME without additional preprocessing. For each entity, we concatenate its name with its textual description to form a descriptive sentence, which is then embedded using GME. Relation embeddings are constructed similarly, by combining relation keywords, the names of the source and target entities, and a textual description. All embeddings are stored in separate dense vector stores by type. 

\noindent\textbf{Graph Retrieval.}
To retrieve relevant knowledge, we adopt a dual-level retrieval strategy~\cite{guo2024lightrag} that targets both entities and relations. Given a user query, we first prompt the MLLM to extract two types of keywords: low-level keywords corresponding to specific entities, and high-level keywords that capture broader concepts. These keywords are then embedded by using the same GME model adopted during indexing.
Both low-level and high-level keywords are combined into a single keyword list and used to query the entity vector store, retrieving the top-$k$ most relevant entities. In parallel, the top-$k$ most relevant relations, along with their associated source and target entities, are retrieved from the relation store. To further enrich the context, each retrieved entity is expanded by incorporating its one-hop neighbors from \(\mathcal{G}^{1}\). The final set of entities and relations serves as input to the downstream reasoning module.

\noindent\textbf{Page Retrieval.} Complementary to graph retrieval, we also perform text-to-page(image) retrieval to capture fine-grained visual and layout cues that may be missed by symbolic representations alone. Given the same input query, we retrieve the top-$m$ relevant document pages by comparing text and image embeddings within the shared vector space.

\subsection{MMKG-augmented Generation}
When combined with visual content and MMKG in a single MLLM prompt, this integration can lead to modality bias. The model often disproportionately focuses on one modality, typically text, while underutilizing the other.
To address this issue, we propose a two-stage answer generation approach that decouples the processing of textual and visual inputs. Given the retrieved subgraph and the relevant page images, the model first generates two intermediate responses in parallel: one based on the symbolic knowledge graph, and the other on the visual content. In the second stage, the MLLM synthesizes a final answer by integrating both intermediate outputs. Full prompt formats for each generation stage are provided in Appendix~\ref{sec:appendix_implementation_details}.

\section{Experiments}
\label{experiments}
In this section, we outline the experimental setups and present the results for our MegaRAG method.

\subsection{Datasets}
\noindent\textbf{Global QA.}
To evaluate the global (book-level) QA capabilities of MegaRAG, we use two document collections: a textual corpus and a multimodal dataset. For the \textbf{textual benchmark}, we adopt the Ultradomain~\cite{qian2024memorag} dataset, which contains 428 college-level textbooks across 18 disciplines; we focus on four representative subsets: Agriculture (2,017,886 tokens), Legal (5,081,069 tokens), Computer Science (2,306,535 tokens) and Mixed-Domain (619,009 tokens). Since no standard benchmark exists for multimodal global QA, we curate a new \textbf{multimodal benchmark} comprising four documents: World History (a world history textbook, 788 pages), Environmental Report (a corporate environmental report slide deck, 422 pages), DLCV (an English lecture slide deck, 1,984 pages), and GenAI (a Chinese lecture slide deck, 594 pages).

\begingroup
\setlength{\tabcolsep}{8pt}     
\renewcommand{\arraystretch}{1.1} 

\begin{table*}
\centering
\resizebox{\linewidth}{!}{%
\begin{tabular}{lcccccccccccc} 
\toprule
 & \multicolumn{3}{c}{\textbf{DLCV}} & \multicolumn{3}{c}{\textbf{World History}} & \multicolumn{3}{c}{\textbf{Environmental Report}} & \multicolumn{3}{c}{\textbf{GenAI}} \\ 
\cmidrule(r){2-4}\cmidrule(r){5-7}\cmidrule(r){8-10}\cmidrule(lr){11-13}
 & NaiveRAG & \textbf{Ours} & Tie & NaiveRAG & \textbf{Ours} & Tie & NaiveRAG & \textbf{Ours} & Tie & NaiveRAG & \textbf{Ours} & Tie \\ 
\hline
Comprehensiveness & 2.4 & \textbf{67.2} & 30.4 & 0.0 & \textbf{81.5} & 18.5 & 0.0 & \textbf{72.8} & 27.2 & 0.0 & \textbf{95.2} & 4.8 \\
Diversity & 6.4 & \textbf{84.8} & 8.8 & 0.0 & \textbf{96.8} & 3.2 & 2.4 & \textbf{92.0} & 5.6 & 0.0 & \textbf{98.4} & 1.6 \\
Empowerment & 11.2 & \textbf{66.4} & 22.4 & 3.2 & \textbf{82.3} & 14.5 & 12.8 & \textbf{64.0} & 23.2 & 0.8 & \textbf{88.0} & 11.2 \\
Overall & 4.8 & \textbf{75.2} & 20.0 & 0.0 & \textbf{89.5} & 10.5 & 1.6 & \textbf{80.0} & 18.4 & 0.0 & \textbf{98.4} & 1.6 \\ 
\cmidrule(lr){2-4}\cmidrule(lr){5-7}\cmidrule(lr){8-10}\cmidrule(lr){11-13}
 & GraphRAG & \textbf{Ours} & Tie & GraphRAG & \textbf{Ours} & Tie & GraphRAG & \textbf{Ours} & Tie & GraphRAG & \textbf{Ours} & Tie \\ 
\hline
Comprehensiveness & 0.0 & \textbf{88.8} & 11.2 & 0.0 & \textbf{92.0} & 8.0 & 0.8 & \textbf{68.8} & 30.4 & 0.0 & \textbf{92.8} & 7.2 \\
Diversity & 3.2 & \textbf{92.8} & 4.0 & 1.6 & \textbf{97.6} & 0.8 & 7.2 & \textbf{81.6} & 11.2 & 0.0 & \textbf{97.6} & 2.4 \\
Empowerment & 1.6 & \textbf{95.2} & 3.2 & 0.0 & \textbf{96.0} & 4.0 & 1.6 & \textbf{93.6} & 4.8 & 0.0 & \textbf{100.0} & 0.0 \\
Overall & 0.0 & \textbf{92.8} & 7.2 & 0.0 & \textbf{93.6} & 6.4 & 0.8 & \textbf{84.8} & 14.4 & 0.0 & \textbf{99.2} & 0.8 \\ 
\cmidrule(lr){2-4}\cmidrule(lr){5-7}\cmidrule(lr){8-10}\cmidrule(lr){11-13}
 & LightRAG & \textbf{Ours} & Tie & LightRAG & \textbf{Ours} & Tie & LightRAG & \textbf{Ours} & Tie & LightRAG & \textbf{Ours} & Tie \\ 
\hline
Comprehensiveness & 0.0 & \textbf{78.4} & 21.6 & 0.0 & \textbf{89.6} & 10.4 & 0.0 & \textbf{80.8} & 19.2 & 0.0 & \textbf{92.0} & 8.0 \\
Diversity & 3.2 & \textbf{90.4} & 6.4 & 0.0 & \textbf{95.2} & 4.8 & 1.6 & \textbf{92.0} & 6.4 & 1.6 & \textbf{92.8} & 5.6 \\
Empowerment & 11.2 & \textbf{74.4} & 14.4 & 3.2 & \textbf{86.4} & 10.4 & 4.8 & \textbf{79.2} & 16.0 & 1.6 & \textbf{91.2} & 7.2 \\
Overall & 0.8 & \textbf{84.8} & 14.4 & 0.0 & \textbf{90.4} & 9.6 & 0.0 & \textbf{90.4} & 9.6 & 0.0 & \textbf{94.4} & 5.6 \\
\bottomrule
\end{tabular}
}
\caption{Performance across four multimodal datasets in terms of win rates (\%).}
\label{tab:main_result_mmdocs}
\end{table*}
\endgroup
\begingroup
\setlength{\tabcolsep}{8pt}     
\renewcommand{\arraystretch}{0.8} 

\begin{table}[t]
\centering
\resizebox{1\linewidth}{!}{%
\begin{tabular}{lccccc}
\toprule
\multirow{2}{*}{Method} & \multirow{2}{*}{SlideVQA (2k)} &
\multicolumn{4}{c}{RealMMBench} \\
\cmidrule(lr){3-6}
& & FinReport & FinSlides & TechReport & TechSlides \\ 
\midrule
NaiveRAG        & 11.34  & 29.66 & 14.64 & 36.63 & 32.94 \\
GraphRAG (L)    & 6.80  & 24.50 & 11.98 & 29.60 & 26.81 \\
GraphRAG (G)    & 5.22  & 10.08 &  3.04 & 15.07 & 16.03 \\
LightRAG        & 27.66  & 31.30 & 13.02 & 42.74 & 31.39 \\
\textbf{MegaRAG} & \textbf{64.85}  & \textbf{39.51} & \textbf{58.37} & \textbf{51.51} & \textbf{60.86} \\
\bottomrule
\end{tabular}}
\label{tab:main_result_realmmbench}
\caption{
Performance on SlideVQA (2k) and RealMMBench datasets in terms of Accuracy (\%). GraphRAG (L) and GraphRAG (G) denote its local and global search modes.}
\label{tab:local-acc}
\end{table}

\endgroup

As these datasets lack manually labeled global questions, we adopt the question generation strategy from GraphRAG \cite{edge2024local} and LightRAG \cite{guo2024lightrag}. For each dataset, we use the document outline as input and prompt an LLM to create five synthetic RAG users, each with a profile describing their background and information needs. Each user is assigned five tasks representing distinct information-seeking goals, and each task is used to generate five questions that require a comprehensive understanding of the full document. This process yields 125 global questions per dataset.

\noindent\textbf{Local QA.} To evaluate local (slide- or page-level) QA, we use two benchmarks: SlideVQA~\cite{yang2018hotpotqa} and RealMMBench~\cite{wasserman2025real}. SlideVQA includes over 52,000 slides and 14,500 questions covering complex reasoning and numerical understanding, but its scale makes full evaluation computationally expensive. Instead, we construct a subset of 2,000 slides, referred to as SlideVQA (2k). RealMMBench assesses retrieval in multimodal RAG settings using visual-rich, table-heavy, and rephrased queries. RealMMBench consists of four sub-datasets: FinReport (2,687 pages), FinSlides (2,280 pages), TechReport (1,674 pages), and TechSlides (1,963 pages).
Additional details are provided in Appendix~\ref{sec:appendix_datasets}.

\subsection{Baselines and Evaluation Metrics}
We compare MegaRAG with standard RAG baselines, including NaiveRAG and two recent KG-based methods, GraphRAG~\cite{edge2024local} and LightRAG~\cite{guo2024lightrag} (details in Appendix~\ref{sec:appendix_baselines_and_evaluation}). For fair comparison, we evaluate all methods on both the multimodal benchmark and the textual benchmark.

\noindent\textbf{Global QA.}
Since global (book-level) questions lack gold answers, we adopt the pairwise LLM-judge protocol from GraphRAG/LightRAG~\cite{edge2024local,guo2024lightrag}. Responses are evaluated by win rate (ties included) on four dimensions: Comprehensiveness, Diversity, Empowerment, and Overall~\cite{guo2024lightrag}.

\noindent\textbf{Local QA.}
For local (page/slide-level) QA with references, an LLM judges semantic equivalence between the generated and ground-truth answers, and we report accuracy. Additional evaluation details are provided in Appendix~\ref{sec:appendix_baselines_and_evaluation}.

\subsection{Implementation Details}
\label{sec:implementation}
To ensure fair comparisons, we standardize the LLM/MLLM and prompting across all methods. Response generation and global question generation use GPT-4o-mini, and evaluation uses GPT-4.1-mini. All baselines (NaiveRAG, GraphRAG, LightRAG) use text-embedding-3-small for textual embeddings. Textual documents are chunked into 1,200 tokens with a 100-token overlap; for GraphRAG and LightRAG, we set gleaning to 1. We use temperature 0 for all tasks.

For multimodal documents, we use MinerU~\cite{wang2024mineru} to extract text, figures, and tables. MegaRAG encodes multimodal embeddings with GME-Qwen2-VL-2B~\cite{zhang2024gme}. We retrieve top-$k$ entities/relations with $k=60$ for graph retrieval and top-$m$ pages with $m=6$ for page retrieval (Section~\ref{sec:indexing_retrival}). For text-only baselines, we keep only extracted text and apply the same textual pipeline.

\subsection{Main Results}

\noindent\textbf{Textual Global QA.}
Table~\ref{tab:main_result_ultradomain} 
shows the results on the UltraDomain benchmark consisting of purely textual documents. As can be seen, across all domains and evaluation dimensions, MegaRAG consistently outperforms the baselines, achieving average win rates of 59.0\% for Comprehensiveness, 71.4\% for Diversity, 74.8\% for Empowerment, and 71.8\% Overall.
 
A key contributor to this performance is MegaRAG’s graph refinement process. Unlike GraphRAG and LightRAG, which employ gleaning per page, a form of local subgraph refinement, MegaRAG doesn't employ gleaning but constructs and refines a global knowledge graph that captures broader contextual relationships between documents. This approach enhances the expressiveness and coverage of the graph, leading to superior performance.

\begingroup
\setlength{\tabcolsep}{8pt}     
\renewcommand{\arraystretch}{1.1} 

\begin{table*}
\centering
\resizebox{\linewidth}{!}{%
\begin{tabular}{lcccccccccccc} 
\toprule
 & \multicolumn{3}{c}{\textbf{DLCV}} & \multicolumn{3}{c}{\textbf{World History}} & \multicolumn{3}{c}{\textbf{Environmental Report}} & \multicolumn{3}{c}{\textbf{GenAI}} \\ 
\cmidrule(l){2-13}
 & VisRAG & \textbf{Ours} & Tie & VisRAG & \textbf{Ours} & Tie & VisRAG & \textbf{Ours} & Tie & VisRAG & \textbf{Ours} & Tie \\ 
\hline
Comprehensiveness & 0.8 & \textbf{35.2} & 64.0 & 4.0 & \textbf{31.2} & 64.8 & 4.0 & \textbf{27.2} & 68.8 & 0.0 & \textbf{87.2} & 12.8 \\
Diversity & 26.4 & \textbf{52.0} & 21.6 & 15.2 & \textbf{64.8} & 20.0 & 15.2 & \textbf{60.8} & 24.0 & 2.4 & \textbf{94.4} & 3.2 \\
Empowerment & 12.0 & \textbf{69.6} & 18.4 & 16.0 & \textbf{56.8} & 27.2 & 32.8 & \textbf{32.0} & 35.2 & 1.6 & \textbf{93.6} & 4.8 \\
Overall & 8.0 & \textbf{53.6} & 38.4 & 4.0 & \textbf{56.0} & 40.0 & 9.6 & \textbf{40.0} & 50.4 & 0.0 & \textbf{96.0} & 4.0 \\ 
\cmidrule(lr){2-4}\cmidrule(lr){5-7}\cmidrule(lr){8-10}\cmidrule(lr){11-13}
 & GME & \textbf{Ours} & Tie & GME & \textbf{Ours} & Tie & GME & \textbf{Ours} & Tie & GME & \textbf{Ours} & Tie \\ 
\hline
Comprehensiveness & 6.4 & \textbf{49.6} & 44.0 & 0.0 & \textbf{72.0} & 28.0 & 2.4 & \textbf{48.8} & 48.8 & 0.8 & \textbf{75.2} & 24.0 \\
Diversity & 19.2 & \textbf{59.2} & 21.6 & 5.6 & \textbf{80.8} & 13.6 & 12.0 & \textbf{75.2} & 12.8 & 4.8 & \textbf{86.4} & 8.8 \\
Empowerment & 23.2 & \textbf{49.6} & 27.2 & 5.6 & \textbf{75.2} & 19.2 & 24.0 & \textbf{50.4} & 25.6 & 6.4 & \textbf{72.0} & 21.6 \\
Overall & 14.4 & \textbf{57.6} & 28.0 & 1.6 & \textbf{78.4} & 20.0 & 5.6 & \textbf{64.0} & 30.4 & 0.8 & \textbf{86.4} & 12.8 \\ 
\cmidrule(lr){2-4}\cmidrule(lr){5-7}\cmidrule(lr){8-10}\cmidrule(lr){11-13}
 & ColQwen & \textbf{Ours} & Tie & ColQwen & \textbf{Ours} & Tie & ColQwen & \textbf{Ours} & Tie & ColQwen & \textbf{Ours} & Tie \\ 
\hline
Comprehensiveness & 2.4 & \textbf{39.5} & 58.1 & 1.6 & \textbf{28.8} & 69.6 & 2.4 & \textbf{37.6} & 60.0 & 0.0 & \textbf{88.8} & 11.2 \\
Diversity & 29.0 & \textbf{57.3} & 13.7 & 14.4 & \textbf{65.6} & 20.0 & 17.6 & \textbf{57.6} & 24.8 & 1.6 & \textbf{94.4} & 4.0 \\
Empowerment & 14.5 & \textbf{62.9} & 22.6 & 16.0 & \textbf{53.6} & 30.4 & 25.6 & \textbf{44.8} & 29.6 & 1.6 & \textbf{90.4} & 8.0 \\
Overall & 6.5 & \textbf{50.8} & 42.7 & 3.2 & \textbf{52.0} & 44.8 & 9.6 & \textbf{50.4} & 40.0 & 0.0 & \textbf{94.4} & 5.6 \\
\bottomrule
\end{tabular}
}
\caption{Pairwise LLM-judge win rates (\%) of MegaRAG against MMRAG baselines on four multimodal datasets.}
\label{tab:appendix_result_mmdocs_mmrag}
\vspace{-5pt}
\end{table*}
\endgroup

\noindent\textbf{Multimodal Global QA.}
A main characteristic of our method is that it can build MMKGs for RAG.
In this experiment, we evaluate our MegaRAG on global QA tasks over multimodal documents. As shown in Table~\ref{tab:main_result_mmdocs}, MegaRAG outperforms all baselines on four visually rich datasets: World History, Environmental Report, DLCV, and GenAI. It achieves average win rates of 83.3\% for Comprehensiveness, 92.7\% for Diversity, 84.7\% for Empowerment, and 89.5\% Overall. The advantage is particularly evident on slide-based datasets such as DLCV and GenAI, where much of the core content is visual rather than textual. 
Compared with NaiveRAG and LightRAG, relying primarily on text, MegaRAG delivers stronger results across all evaluation dimensions. These gains stem from MegaRAG’s ability to build KGs that jointly encode textual information and visual cues. 

Although all baselines in this comparison are text-only models, our ablation study, Section~\ref{sec:ablation_study}, further demonstrates that removing MMKG from MegaRAG leads to a substantial performance drop. 
Since our MegaRAG reduces to an MMRAG approach when its KG components are removed, this suggests that even vision-capable retrieval methods of MMRAG would struggle to match MegaRAG without multimodal global knowledge integration.

\noindent\textbf{Multimodal Local QA.}
Table~\ref{tab:local-acc} shows the accuracy results on SlideVQA (2k) and the four RealMMBench subsets. Across all five test sets, MegaRAG performs more favorably. On SlideVQA (2k), which focuses on fine-grained slide-level reasoning, MegaRAG achieves 64.85\% accuracy, higher than double the score of the strongest baseline. 
Similar trends are observed in RealMMBench. On FinSlides and TechSlides, which feature highly visual and table slide content, MegaRAG achieves 58.37\% and 60.86\%, outperforming the best baseline by 45 and 29 percents, respectively. Even in the more text-heavy FinReport and TechReport subsets, MegaRAG leads with 39.51\% and 51.51\%, surpassing LightRAG by 8 to 9\%.

\subsection{Comparison with Multimodal RAG Methods}
\label{sec:mmrag_comparison}
Table~\ref{tab:appendix_result_mmdocs_mmrag} presents pairwise LLM-judge win rates between MegaRAG and three strong multimodal RAG methods, namely VisRAG, GME, and ColQwen, on four multimodal documents.

MegaRAG consistently outperforms all three baselines across all datasets and evaluation dimensions, often by a substantial margin. While VisRAG, GME, and ColQwen mainly rely on multimodal retrieval over raw page images or unified multimodal embeddings, they do not explicitly model corpus-level structure or cross-page relations. By contrast, MegaRAG constructs multimodal knowledge graphs automatically, enabling structured abstraction and cross-page, multi-hop reasoning. This design yields particularly strong improvements in \textit{Comprehensiveness} and \textit{Empowerment}, where capturing dispersed evidence are important.

\begin{table}
\centering
\resizebox{\linewidth}{!}{%
\begin{tabular}{lcccccc} 
\toprule
 & \multicolumn{3}{c}{\textbf{DLCV}} & \multicolumn{3}{c}{\textbf{World History}} \\ 
\cmidrule(l){2-7}
 & A1 & \textbf{MegaRAG} & Tie & A1 & \textbf{MegaRAG} & Tie \\ 
\hline
Comprehensiveness & 6.4 & \textbf{49.6} & 44.0 & 0.0 & \textbf{72.0} & 28.0 \\
Diversity & 19.2 & \textbf{59.2} & 21.6 & 5.6 & \textbf{80.8} & 13.6 \\
Empowerment & 23.2 & \textbf{49.6} & 27.2 & 5.6 & \textbf{75.2} & 19.2 \\
Overall & 14.4 & \textbf{57.6} & 28.0 & 1.6 & \textbf{78.4} & 20.0 \\ 
\cmidrule(lr){2-4}\cmidrule(lr){5-7}
 & A2 & \textbf{MegaRAG} & Tie & A2 & \textbf{MegaRAG} & Tie \\ 
\hline
Comprehensiveness & 0.0 & \textbf{100.0} & 0.0 & 0.8 & \textbf{88.0} & 11.2 \\
Diversity & 0.0 & \textbf{100.0} & 0.0 & 0.8 & \textbf{96.0} & 3.2 \\
Empowerment & 0.0 & \textbf{100.0} & 0.0 & 1.6 & \textbf{86.4} & 12.0 \\
Overall & 0.0 & \textbf{100.0} & 0.0 & 0.8 & \textbf{91.2} & 8.0 \\ 
\cmidrule(lr){2-4}\cmidrule(lr){5-7}
 & A3 & \textbf{MegaRAG} & Tie & A3 & \textbf{MegaRAG} & Tie \\ 
\hline
Comprehensiveness & 0.8 & \textbf{52.8} & 46.4 & 0.0 & \textbf{67.2} & 32.8 \\
Diversity & 12.0 & \textbf{72.8} & 15.2 & 8.0 & \textbf{79.2} & 12.8 \\
Empowerment & 5.6 & \textbf{70.4} & 24.0 & 4.0 & \textbf{77.6} & 18.4 \\
Overall & 1.6 & \textbf{61.6} & 36.8 & 0.8 & \textbf{75.2} & 24.0 \\
\bottomrule
\end{tabular}
}
\caption{Ablations on multimodal datasets (win rates, \%). A1: text-only graph; A2: page-only retrieval; A3: single-pass (w/o two-stage) generation.}

\label{tab:main_result_mmdocs_ablations}
\vspace{-5pt}
\end{table}

\subsection{Ablation Study}
\label{sec:ablation_study}

We study the impact of MegaRAG's core components by ablating modules in three stages: MMKG construction, retrieval, and answer generation. Table~\ref{tab:main_result_mmdocs_ablations} reports the ablation results. (A1) removes all visual inputs (figures, tables, and page images) during graph construction, yielding a text-only graph. (A2) disables MMKG retrieval and uses only page retrieval. (A3) replaces our two-stage generation with a single-pass setup that jointly conditions on the retrieved subgraph and page images.

\noindent\textbf{(A1) Text-only graph construction.}
Removing visual cues consistently degrades performance. Without visual entities and cross-modal links, the graph loses critical grounding signals, which is particularly harmful for visually intensive documents. This confirms the necessity of incorporating visual elements when building the MMKG.

\noindent\textbf{(A2) Disable MMKG retrieval.}
This variant yields the largest drop. Relying solely on page retrieval substantially weakens evidence aggregation and multi-hop reasoning, and MegaRAG strongly outperforms this setting across evaluation dimensions. The results indicate that structured retrieval over the MMKG is the primary driver of our gains.

\noindent\textbf{(A3) Remove two-stage answer generation.}
Using a single-pass generation causes moderate but consistent regressions. The largest decreases are typically observed in Diversity and Empowerment, suggesting that decoupling graph-based reasoning from visual grounding before fusion helps produce richer and more informative answers.

Overall, MMKG-based retrieval (A2) is the most critical component, while visual inputs in construction (A1) and the two-stage generation strategy (A3) provide complementary improvements.

\begin{table}[t]
\centering
\setlength{\tabcolsep}{4pt}
\resizebox{\columnwidth}{!}{%
\begin{tabular}{ccrcccc} 
\toprule
\multirow{2}{*}{\begin{tabular}[c]{@{}c@{}}\textbf{Dataset}\\\textbf{(Pages)}\end{tabular}} & \multirow{2}{*}{\begin{tabular}[c]{@{}c@{}}\textbf{Input Tokens}\\\textbf{\textbf{(Visual / Text)}}\end{tabular}} & \multicolumn{1}{c}{\multirow{2}{*}{\textbf{Metric}}} & \multirow{2}{*}{\begin{tabular}[c]{@{}c@{}}\textbf{GraphRAG}\\\textbf{(Total)}\end{tabular}} & \multicolumn{3}{c}{\textbf{MegaRAG}} \\ 
\cmidrule(lr){5-7}
 &  & \multicolumn{1}{c}{} &  & Init. & Ref. & \begin{tabular}[c]{@{}c@{}}Total \\(Init. + Ref.)\end{tabular} \\ 
\hline
\multirow{2}{*}{World History~(788)} & \multirow{2}{*}{17.0M / 0.4M} & \textit{Time (min)} & 23.0 & 19.0 & 12.0 & 31.0 \\
 &  & \textit{KG Tokens (M)} & 1.2 & 22.9 & 15.3 & 38.2 \\ 
\hline
\multirow{2}{*}{Env. eport~(422)} & \multirow{2}{*}{25.0M / 0.2M} & \textit{Time (min)} & 10.0 & 8.5 & 5.5 & 14.0 \\
 &  & \textit{KG Tokens (M)} & 0.6 & 32.6 & 21.7 & 54.3 \\
\bottomrule
\end{tabular}
}
\caption{MMKG construction costs, where \textit{Init.} and \textit{Ref.} denote the Initial and Refinement stages.}
\label{tab:indexing_costs}
\vspace{-10pt}
\end{table}

\subsection{Computational Overhead}
\label{sec:computational_overhead}

We analyze the computational overhead of MegaRAG in terms of both indexing cost and inference latency.

\noindent\textbf{Indexing Cost.}
MegaRAG constructs the MMKG in a one-time offline process consisting of an \textit{Initial Stage} and a \textit{Refinement Stage}. Table~\ref{tab:indexing_costs} reports the indexing costs on two representative datasets.

Importantly, graph construction is a one-time offline process; the resulting MMKG can be reused for subsequent queries without rebuilding the graph. Compared to text-only GraphRAG, MegaRAG incurs higher token usage and indexing latency (approximately $1.4\times$) due to processing image content (17M to 25M visual tokens). This additional cost enables MegaRAG to extract visual entities (e.g., 473 to 538 figures) that text-only methods cannot capture and to recover global dependencies missed by single-pass approaches.

\noindent\textbf{Inference Latency.}
MegaRAG adopts a two-stage generation strategy. We measure end-to-end latency on the \textit{Environmental Report} dataset using a single RTX~3090 (24GB); the latency can be further reduced with more powerful GPUs. MMKG retrieval and page retrieval run concurrently ($\sim$1.0s in total). In Stage~1, the MLLM generates answers from the retrieved subgraph ($\sim$20.5s) and from page images ($\sim$26s) in parallel. In Stage~2, a fusion step integrates both outputs ($\sim$16s), resulting in a total latency of $\sim$42s per question.

For comparison, GME, a strong MMRAG method, completes retrieval ($\sim$0.4s) and generation ($\sim$26s), resulting in a total latency of $\sim$26.4s. Although MegaRAG introduces an additional $\sim$15.6s latency, it produces MMKG-augmented answers that combine textual and visual evidence for stronger cross-page and multi-hop reasoning.

\section{Conclusion}
\label{conclusion}
In this paper, we introduced MegaRAG, a novel KG-based RAG method that leverages MLLMs to automatically construct MMKGs. MegaRAG improves MLLMs' capabilities over complex, long-form documents by combining textual and visual information into a unified graph representation and refining it through iterative updates. MegaRAG needs no fine-tuning and is easy to use. To reduce modality bias, we adopt a two-stage answer generation process that separately reasons over textual and visual evidence before integrating the results, enabling more comprehensive and balanced responses. Through evaluations on both global and local QA tasks across textual and multimodal datasets, MegaRAG consistently outperforms other competitive RAG approaches.
Our work highlights a promising new direction for scalable and interpretable multimodal reasoning in RAG systems.

\section*{Limitations}
Our experiments are primarily conducted on knowledge graphs constructed within a single book or report, and question answering across multiple books or reports remains limited. In addition, due to the high computational and preprocessing costs associated with figures and images, the scale of our multimodal dataset is still relatively limited.

Another limitation lies in our current treatment of visual content: each figure is represented as a single entity, which may overlook finer-grained elements or objects that are individually meaningful and worth representing as distinct entities. Moreover, since the MMKG is automatically constructed by MLLMs, hallucinations in entity or relation extraction may propagate to downstream reasoning. Understanding such hallucination behaviors and developing robust verification mechanisms are important directions for future work.

\section*{Acknowledgements}
This work was supported in part by NSTC under Grants 112-2221-E-002-132-MY3 and 113-2634-F-002-003, NTU under Grant 114L900902, and the Esun-NTU collaboration project under Grant 114HZA3S002. We thank the National Center for High performance Computing (NCHC) of NARLabs in Taiwan for providing computational and storage resources.

\section*{Ethical Considerations}
All data used in this work are publicly available documents. No human subjects or user-generated data are involved, and we do not collect or infer personally identifying information beyond what is already contained in the original sources. Our approach relies on LLMs, which may occasionally produce incorrect or biased text. In our setting, LLM outputs are primarily used to assist knowledge graph construction, while final answers are grounded in retrieved subgraphs and source pages and can be verified against the underlying documents, which helps limit potential errors.

\bibliography{custom}
\clearpage

\appendix
\label{sec:appendix}

\section*{Appendix}

In the Appendix, we present the Datasets, Implementation Details, and Baselines \& Evaluations in Appendices~\ref{sec:appendix_datasets}, \ref{sec:appendix_implementation_details}, and \ref{sec:appendix_baselines_and_evaluation}, respectively.

\begingroup
\setlength{\tabcolsep}{2pt}     
\renewcommand{\arraystretch}{0.8} 

\begin{table}[htbp]
\centering
\setlength{\extrarowheight}{0pt}
\addtolength{\extrarowheight}{\aboverulesep}
\addtolength{\extrarowheight}{\belowrulesep}
\setlength{\aboverulesep}{0pt}
\setlength{\belowrulesep}{0pt}
\resizebox{\linewidth}{!}{%
\begin{tabular}{lrrrrr} 
\toprule
\multicolumn{1}{c}{\textbf{Dataset}} & \multicolumn{1}{c}{\textbf{Documents}} & \multicolumn{1}{c}{\textbf{Pages}} & \multicolumn{1}{c}{\textbf{Figures}} & \multicolumn{1}{c}{\textbf{Tables}} & \multicolumn{1}{c}{\textbf{Text Tokens}} \\ 
\midrule
\multicolumn{6}{l}{{\cellcolor[rgb]{0.929,0.929,0.929}}\textbf{\textit{Ultradomain}}} \\
Agriculture & 12 & \multicolumn{1}{c}{-} & \multicolumn{1}{c}{-} & \multicolumn{1}{c}{-} & 2,017,886 \\
Computer Science (CS) & 10 & \multicolumn{1}{c}{-} & \multicolumn{1}{c}{-} & \multicolumn{1}{c}{-} & 2,306,535 \\
Legal & 94 & \multicolumn{1}{c}{-} & \multicolumn{1}{c}{-} & \multicolumn{1}{c}{-} & 5,081,069 \\
Mix & 61 & \multicolumn{1}{c}{-} & \multicolumn{1}{c}{-} & \multicolumn{1}{c}{-} & 619,009 \\ 
\hline
\multicolumn{6}{l}{{\cellcolor[rgb]{0.929,0.929,0.929}}\textbf{\textit{Multimodal Documents}}} \\
DLCV & 18 & 1,984 & 2,018 & 75 & 136,032 \\
Environmental Report & 5 & 422 & 416 & 122 & 229,014 \\
GenAI & 20 & 594 & 686 & 33 & 55,913 \\
World History & 1 & 788 & 468 & 5 & 441,764 \\ 
\hline
\multicolumn{6}{l}{{\cellcolor[rgb]{0.929,0.929,0.929}}\textbf{\textit{SlideVQA}}} \\
SlideVQA (2k) & 100 & 2,000 & 1,581 & 139 & 119,776 \\ 
\hline
\multicolumn{6}{l}{{\cellcolor[rgb]{0.929,0.929,0.929}}\textbf{\textit{RealMMBench}}} \\
FinReport & 19 & 2,687 & 411 & 2,963 & 1,583,640 \\
FinSlides & 65 & 2,280 & 730 & 1,842 & 123,891 \\
TechReport & 17 & 1,674 & 928 & 337 & 535,415 \\
TechSlides & 62 & 1,963 & 2,254 & 119 & 138,766 \\
\bottomrule
\end{tabular}
}
\caption{Datasets statistics used in our experiments. 
The Ultradomain benchmark is purely textual documents; hence, entries for pages, figures, and tables are marked with a dash (–) to indicate not applicable.}
\label{tab:ultradomain_stats}
\end{table}

\endgroup

\section{Datasets}
\label{sec:appendix_datasets}

We provide an overview of the datasets in our experiments and dataset statistics in Table~\ref{tab:ultradomain_stats}.

\subsection{Dataset Statistics}

The Ultradomain benchmark~\citep{qian2024memorag} comprises 428 college-level textbooks spanning 18 academic disciplines. For this study, we focus on the four representative subsets:

\noindent\textbf{Agriculture dataset.}  
Consisting of 12 textbooks and 2.02 million text tokens, this subset covers topics such as beekeeping, hive management, crop cultivation, and disease prevention in modern agriculture.
\textbf{Computer Science (CS) dataset.}  
Containing 10 textbooks and 2.31 million tokens, the CS subset emphasizes key topics in algorithms, data structures, artificial intelligence, machine learning, and real-time data analytics.
\textbf{Legal dataset.}  
Comprising 94 textbooks and totaling 5.08 million tokens. It spans a wide range of legal topics, including corporate restructuring, regulatory compliance, financial governance, and case law analysis.
\textbf{Mixed-Domain (Mix) dataset.}  
A diverse collection of 61 textbooks totaling 620,000 tokens. This subset includes literary works, philosophical essays, biographies, and cultural-historical studies.

The global QA multimodal datasets are derived from publicly available documents:

\noindent\textbf{Deep Learning for Computer Vision (DLCV) dataset.}  
Comprising 18 slide decks, this dataset\footnote{\scriptsize{\url{https://cs231n.stanford.edu/slides/2024/}}} includes 1,984 pages, 2,018 figures, 75 tables, and 136,000 tokens. The content is drawn from a deep learning and computer vision course, covering image classification, object detection, and societal impacts of AI.
\textbf{Environmental Report dataset.}  
Consisting of 5 corporate sustainability reports, this dataset includes 422 pages, 416 figures, 122 tables, and 229,000 tokens. It documents environmental strategies from Google\footnote{\scriptsize{\url{https://sustainability.google/reports/google-2024-environmental-report/}}},  
Apple\footnote{\scriptsize{\url{https://www.apple.com/environment/pdf/Apple_Environmental_Progress_Report_2024.pdf}}},  
Microsoft\footnote{\scriptsize{\url{https://www.microsoft.com/en-us/corporate-responsibility/sustainability/report}}},  
Meta\footnote{\scriptsize{\url{https://sustainability.atmeta.com/2024-sustainability-report/}}},  
and NVIDIA\footnote{\scriptsize{\url{https://www.nvidia.com/en-us/sustainability/}}} (FY24 Sustainability Report), including goals for carbon reduction and renewable energy.
\textbf{Generative AI (GenAI) dataset.}  
This dataset comprises 20 lecture slide decks\footnote{\scriptsize{\url{https://speech.ee.ntu.edu.tw/~hylee/genai/2024-spring.php}}} (in Chinese), with 594 pages, 686 figures, 33 tables, and 55,900 tokens. Topics focus on generative AI, including transformer architectures, generation techniques, cross-modal applications, and ethical considerations in large-scale AI systems.
\textbf{World History dataset.}  
A textbook\footnote{\scriptsize{\url{https://open.umn.edu/opentextbooks/textbooks/1418}}} comprising 788 pages, 468 figures, 5 tables, and 442,000 tokens. It traces global developments from prehistory to 1500 CE, covering early civilizations, empires, religious movements, and intercultural exchanges.

\noindent\textbf{SlideVQA (2k).}
SlideVQA~\citep{tanaka2023slidevqa} includes over 52,000 slides and 14,500 questions covering complex reasoning and numerical understanding, but its scale makes full evaluation computationally expensive. Instead, we construct a subset of SlideVQA, which consists of 2,000 educational slides, featuring 1,581 figures, 139 tables, and 120,000 tokens.

The RealMMBench~\cite{wasserman2025real} is designed to evaluate retrieval performance in realistic multi-modal RAG scenarios, and contains four subsets:

\noindent\textbf{FinReport.}  
This subset includes 19 long-form table-heavy financial reports from IBM, totaling 2,687 pages, 411 figures, 2,963 tables, and 1.58 million tokens. 
\textbf{FinSlides.}  
Comprising 65 corporate financial slide decks, this subset spans 2,280 pages, 730 figures, 1,842 tables, and 124,000 tokens. It presents a more visual but still data-rich format for financial information, including quarterly earnings briefings, strategic outlooks, and KPI dashboards.
\textbf{TechReport.}  
This collection includes 17 technical reports with 1,674 pages, 928 figures, 337 tables, and 535,000 tokens. Documents are sourced from specialized domains such as enterprise hardware and storage systems.
\textbf{TechSlides.}  
Featuring 62 technical presentation slide decks, this subset comprises 1,963 pages, 2,254 figures, 119 tables, and 139,000 tokens. It has the highest figure density across RealMMBench, which conveys technical concepts through diagrams and flowcharts.

\subsection{Global Question Generation}
\begin{figure*}
    \centering
    \includegraphics[width=0.7\linewidth]{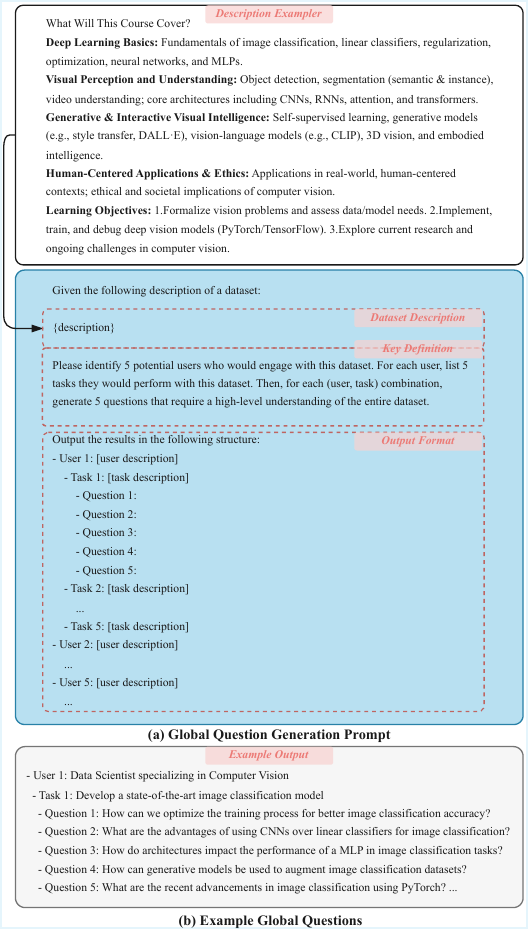}
    \caption{(a) Prompt used for global question generation. (b) Example global questions.}
    \label{fig:global_qg}
\end{figure*}

To generate global questions, we utilize the prompt shown in Figure~\ref{fig:global_qg}. This prompt guides the MLLM (GPT-4o-mini) to first identify representative user profiles and their associated tasks, then generate questions that require a comprehensive understanding of the dataset. 

\section{Implementation Details}
\label{sec:appendix_implementation_details}
\subsection{Prompts Used in MegaRAG}

\begin{figure*}
    \centering
    \includegraphics[width=0.7\linewidth]{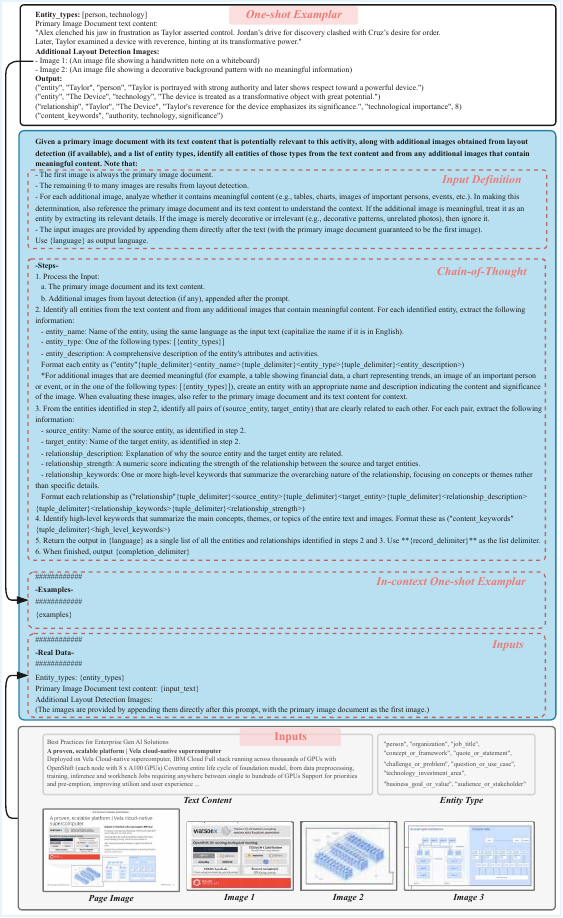}
    \caption{Prompt for extracting entities and relations during the initial construction of the MMKG.}
    \label{fig:MMKG_init}
\end{figure*}

\begin{figure*}
    \centering
    \includegraphics[width=0.7\linewidth]{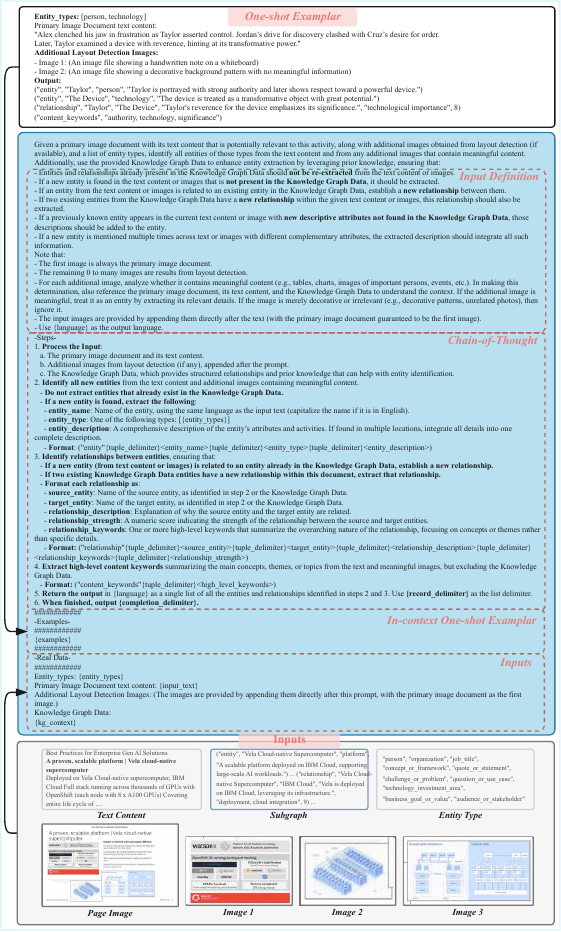}
    \caption{Prompt for MMKG refinement stage.}
    \label{fig:MMKG_refine}
\end{figure*}

\noindent\textbf{MMKG construction.}
For MMKG construction in Section 3.1, we use prompts to guide GPT-4o-mini in extracting structured knowledge from multimodal document inputs. The prompt used in the initial graph construction stage is shown in Figure~\ref{fig:MMKG_init}. 
For graph refinement, we employ a separate prompt designed to identify missing or implicit connections. This prompt, illustrated in Figure~\ref{fig:MMKG_refine}.

\begin{figure*}
    \centering
    \includegraphics[width=0.5\linewidth]{figures_appendix/MMKG_2stage_response__cropped.pdf}
    \caption{Prompts for MMKG-augmented answer generation. (a) Generates an intermediate answer from the retrieved pages. (b) Generates an intermediate answer from the retrieved MMKG subgraph. (c) The final answer is produced by combining both intermediate responses.}
    \label{fig:MMKG_2stage}
\end{figure*}

\noindent\textbf{MMKG-augmented Answer Generation.}
For MMKG-augmented answer generation (Section 3.3), we adopt a two-stage prompting strategy. In the first stage, GPT-4o-mini is guided to generate intermediate answers separately: one based on the visual page (Figure~\ref{fig:MMKG_2stage}(a)) and another based on the retrieved subgraph (Figure~\ref{fig:MMKG_2stage}(b)). In the second stage, a follow-up prompt combines these intermediate responses to produce the final answer (Figure~\ref{fig:MMKG_2stage}(c)).

\subsection{Retrieval and Generation Details}

MegaRAG leverages the General Multimodal Embedder (GME)~\citep{zhang2024gme} to encode entities, relations, and page images within a unified embedding space. GME is built upon the Qwen2-VL architecture, a MLLM capable of processing text, images, or combined text–image inputs. It supports a broad range of retrieval tasks, including single-modality retrieval (e.g., text-to-text, image-to-image), cross-modality retrieval (e.g., text-to-image, image-to-text), and fused-modality retrieval (e.g., text with image to text with image). To generate embeddings, GME uses the final hidden state of the last token as the representation of the input. GME’s strength lies in its flexibility and generalization capability, making it well-suited for MegaRAG, which requires seamless integration of both text-to-text and text-to-page (image) retrieval tasks.

\noindent\textbf{GME Encoding Time.} 
In our pipeline, the GME-Qwen2-VL-2B encoder is executed locally to process both text and image inputs. All encoding is performed on a single NVIDIA RTX 3090 GPU with 24GB of VRAM. Due to memory constraints, we limit GME to encoding two page images concurrently, with an average processing time of approximately 0.97 seconds per image. 

During graph retrieval in the MMKG refinement stage, as described in Section 3.1, we retrieve the top 120 entities and relations from the initial MMKG and concatenate them into a single string (as illustrated in Figure~\ref{fig:MMKG_refine}, subgraph). We then truncate this string to a maximum of 32,000 tokens. The truncated string is then used to prompt the MLLM to identify missing entity-relation pairs that were not captured in the initial stage. We experimented with both larger and smaller retrieval sizes and found that retrieving 120 entities and relations provides the best balance between global coverage of the MMKG and input length constraints. 
\newpage

\section{Baselines and Evaluation}
\label{sec:appendix_baselines_and_evaluation}
\begingroup
\setlength{\tabcolsep}{8pt}     
\renewcommand{\arraystretch}{1.1} 

\begin{table*}
\centering
\resizebox{\linewidth}{!}{%
\begin{tabular}{lcccccccccccc} 
\toprule
 & \multicolumn{3}{c}{\textbf{DLCV}} & \multicolumn{3}{c}{\textbf{World History}} & \multicolumn{3}{c}{\textbf{Environmental Report}} & \multicolumn{3}{c}{\textbf{GenAI}} \\ 
\cmidrule(r){2-4}\cmidrule(lr){5-7}\cmidrule(lr){8-10}\cmidrule(r){11-13}
 & 4o-mini & \textbf{MegaRAG} & Tie & 4o-mini & \textbf{MegaRAG} & Tie & 4o-mini & \textbf{MegaRAG} & Tie & 4o-mini & \textbf{MegaRAG} & Tie \\ 
\hline
Comprehensiveness & 0 & \textbf{94.4} & 5.6 & 0 & \textbf{98.4} & 1.6 & 0 & \textbf{96.8} & 3.2 & 0 & \textbf{99.2} & 0.8 \\
Diversity & 0 & \textbf{95.2} & 4.8 & 0 & \textbf{99.2} & 0.8 & 3.2 & \textbf{92.8} & 4 & 0.8 & \textbf{97.6} & 1.6 \\
Empowerment & 7.2 & \textbf{78.4} & 14.4 & 0 & \textbf{93.6} & 6.4 & 1.6 & \textbf{90.4} & 8 & 1.6 & \textbf{95.2} & 3.2 \\
Overall & 0 & \textbf{96.0} & 4 & 0 & \textbf{99.2} & 0.8 & 2.4 & \textbf{97.6} & 2.4 & 0 & \textbf{99.2} & 0.8 \\
\bottomrule
\end{tabular}
}
\caption{Compare MegaRAG with using 
only GPT-4o-mini in terms of win rates (\%).}
\label{tab:no_retrieval}
\end{table*}
\endgroup

\subsection{Baselines}
We evaluate MegaRAG against two widely used graph-based RAG baselines: GraphRAG and LightRAG, as well as a commonly adopted non-graph baseline, NaiveRAG. To ensure a fair comparison, we set the generation temperature to 0 across all models. Below, we provide a detailed overview of each method along with its specific settings for reference.

\noindent\textbf{NaiveRAG.} Serving as a standard baseline among RAG systems, NaiveRAG divides the input document into multiple text chunks, which are then encoded into a vector space using text embeddings. At query time, relevant chunks are retrieved based on the similarity between their embeddings and the query representation.

\noindent\textbf{GraphRAG.} GraphRAG begins by segmenting the input text into chunks and extracting entities and relationships to construct a graph. This graph is subsequently partitioned into communities at multiple levels. During retrieval, GraphRAG identifies entities mentioned in the query and synthesizes answers by referencing summaries of the corresponding communities. Compared to traditional RAG approaches, GraphRAG offers a more structured and high-level understanding of the document.

\noindent\textbf{LightRAG.} LightRAG is a variant of GraphRAG. It is designed to reduce computational overhead while enhancing retrieval quality through a dual-level retrieval mechanism. This design improves both efficiency and effectiveness, offering a better balance between performance and resource usage compared to GraphRAG.

\subsection{Evaluation}

\noindent\textbf{Global QA.}
To evaluate model performance on global (book-level) questions, where no gold-standard answers are available, we conduct pairwise comparative evaluations between MegaRAG and baseline models. Responses are assessed along three qualitative dimensions: \textbf{Comprehensiveness}, \textbf{Diversity}, and \textbf{Empowerment}, as well as an \textbf{overall} rating that reflects performance across all criteria.

\begin{figure*}
    \centering
    \includegraphics[width=0.7\linewidth]{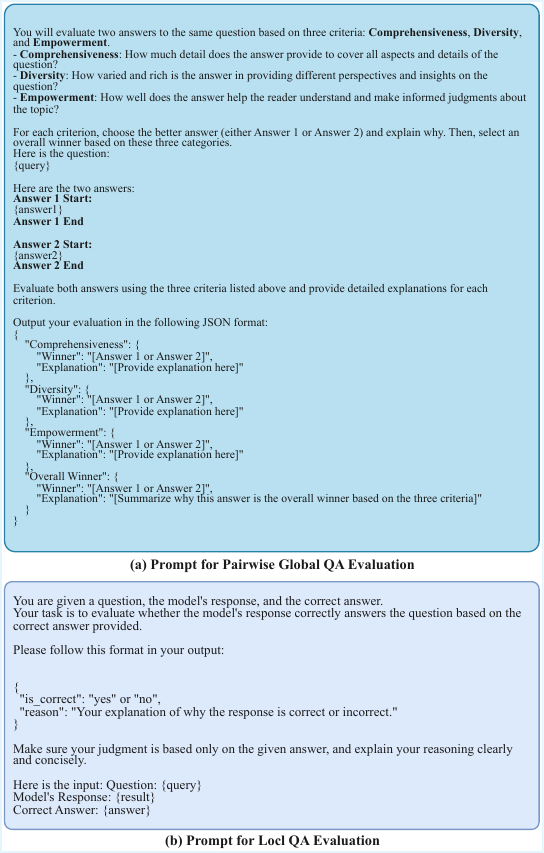}
    \caption{Overview of the global and local QA evaluation prompts.}
    \label{fig:QA_evaluation}
\end{figure*}

Each evaluation instance presents a question alongside two competing answers, one from a baseline model and one from MegaRAG. We employ GPT-4.1-mini as the evaluator to compare the two responses, select a winner for each dimension, and provide brief justifications. Comprehensiveness measures how thoroughly the answer addresses all aspects of the question. Diversity evaluates the richness and variety of perspectives presented. Empowerment assesses how effectively the answer enhances user understanding and supports informed decision-making.
The full evaluation prompt used in this process is shown in Figure~\ref{fig:QA_evaluation} (a).

\noindent\textbf{Local QA.}
For local (slide- or page-level) QA, where reference answers are available, we use GPT-4.1-mini to assess answer correctness. Each instance includes a question, the model’s response, \textbf{}and the corresponding ground truth. The LLM judge evaluates whether the response is semantically consistent with the reference, regardless of surface phrasing. The output is a binary label (\texttt{yes} or \texttt{no}) accompanied by a brief explanation. Accuracy is calculated as the proportion of responses judged correct. The evaluation prompt is shown in Figure~\ref{fig:QA_evaluation} (b).

\begin{table*}[t]
\centering
\footnotesize
\setlength{\tabcolsep}{4pt}
\resizebox{\textwidth}{!}{%
\begin{tabular}{llcccc} 
\toprule
\textbf{Method vs. MegaRAG} & \textbf{Metric} & \textbf{DLCV} & \textbf{World History} & \textbf{Env. eport} & \textbf{GenAI} \\ 
\midrule
\multirow{4}{*}{NaiveRAG} & \textit{Comprehensiveness} & \begin{tabular}[c]{@{}c@{}}2.4/67.2/30.4\\\textbf{(1.9/52.7/45.4)}\end{tabular} & \begin{tabular}[c]{@{}c@{}}0.0/81.5/18.5\\\textbf{(0.0/66.5/33.5)}\end{tabular} & \begin{tabular}[c]{@{}c@{}}0.0/72.8/27.2\\\textbf{(0.0/57.8/42.2)}\end{tabular} & \begin{tabular}[c]{@{}c@{}}0.0/95.2/4.8\\\textbf{(0.0/80.2/19.8)}\end{tabular} \\
 & \textit{Diversity} & \begin{tabular}[c]{@{}c@{}}6.4/84.8/8.8\\\textbf{(5.3/70.9/23.8)}\end{tabular} & \begin{tabular}[c]{@{}c@{}}0.0/96.8/3.2\\\textbf{(0.0/81.8/18.2)}\end{tabular} & \begin{tabular}[c]{@{}c@{}}2.4/92.0/5.6\\\textbf{(1.9/77.5/20.6)}\end{tabular} & \begin{tabular}[c]{@{}c@{}}0.0/98.4/1.6\\\textbf{(0.0/83.4/16.6)}\end{tabular} \\
 & \textit{Empowerment} & \begin{tabular}[c]{@{}c@{}}11.2/66.4/22.4\\\textbf{(9.0/53.6/37.4)}\end{tabular} & \begin{tabular}[c]{@{}c@{}}3.2/82.3/14.5\\\textbf{(2.6/67.9/29.5)}\end{tabular} & \begin{tabular}[c]{@{}c@{}}12.8/64.0/23.2\\\textbf{(10.2/51.6/38.2)}\end{tabular} & \begin{tabular}[c]{@{}c@{}}0.8/88.0/11.2\\\textbf{(0.6/73.2/26.2)}\end{tabular} \\
 & \textit{Overall} & \begin{tabular}[c]{@{}c@{}}4.8/75.2/20.0\\\textbf{(3.9/61.1/35.0)}\end{tabular} & \begin{tabular}[c]{@{}c@{}}0.0/89.5/10.5\\\textbf{(0.0/74.5/25.5)}\end{tabular} & \begin{tabular}[c]{@{}c@{}}1.6/80.0/18.4\\\textbf{(1.3/65.3/33.4)}\end{tabular} & \begin{tabular}[c]{@{}c@{}}0.0/98.4/1.6\\\textbf{(0.0/83.4/16.6)}\end{tabular} \\ 
\midrule
\multirow{4}{*}{GraphRAG} & \textit{Comprehensiveness} & \begin{tabular}[c]{@{}c@{}}0.0/88.8/11.2\\\textbf{(0.0/73.8/26.2)}\end{tabular} & \begin{tabular}[c]{@{}c@{}}0.0/92.0/8.0\\\textbf{(0.0/77.0/23.0)}\end{tabular} & \begin{tabular}[c]{@{}c@{}}0.8/68.8/30.4\\\textbf{(0.6/54.0/45.4)}\end{tabular} & \begin{tabular}[c]{@{}c@{}}0.0/92.8/7.2\\\textbf{(0.0/77.8/22.2)}\end{tabular} \\
 & \textit{Diversity} & \begin{tabular}[c]{@{}c@{}}3.2/92.8/4.0\\\textbf{(2.7/78.3/19.0)}\end{tabular} & \begin{tabular}[c]{@{}c@{}}1.6/97.6/0.8\\\textbf{(1.4/82.8/15.8)}\end{tabular} & \begin{tabular}[c]{@{}c@{}}7.2/81.6/11.2\\\textbf{(6.0/67.8/26.2)}\end{tabular} & \begin{tabular}[c]{@{}c@{}}0.0/97.6/2.4\\\textbf{(0.0/82.6/17.4)}\end{tabular} \\
 & \textit{Empowerment} & \begin{tabular}[c]{@{}c@{}}1.6/95.2/3.2\\\textbf{(1.3/80.5/18.2)}\end{tabular} & \begin{tabular}[c]{@{}c@{}}0.0/81.0/19.0\\\textbf{(0.0/81.0/19.0)}\end{tabular} & \begin{tabular}[c]{@{}c@{}}1.6/93.6/4.8\\\textbf{(1.3/78.9/19.8)}\end{tabular} & \begin{tabular}[c]{@{}c@{}}0.0/100.0/0.0\\\textbf{(0.0/85.0/15.0)}\end{tabular} \\
 & \textit{Overall} & \begin{tabular}[c]{@{}c@{}}0.0/92.8/7.2\\\textbf{(0.0/77.8/22.2)}\end{tabular} & \begin{tabular}[c]{@{}c@{}}0.0/93.6/6.4\\\textbf{(0.0/78.6/21.4)}\end{tabular} & \begin{tabular}[c]{@{}c@{}}0.8/84.8/14.4\\\textbf{(0.7/69.9/29.4)}\end{tabular} & \begin{tabular}[c]{@{}c@{}}0.0/99.2/0.8\\\textbf{(0.0/84.2/15.8)}\end{tabular} \\ 
\midrule
\multirow{4}{*}{LightRAG} & \textit{Comprehensiveness} & \begin{tabular}[c]{@{}c@{}}0.0/78.4/21.6\\\textbf{(0.0/63.4/36.6)}\end{tabular} & \begin{tabular}[c]{@{}c@{}}0.0/89.6/10.4\\\textbf{(0.0/74.6/25.4)}\end{tabular} & \begin{tabular}[c]{@{}c@{}}0.0/80.8/19.2\\\textbf{(0.0/65.8/34.2)}\end{tabular} & \begin{tabular}[c]{@{}c@{}}0.0/92.0/8.0\\\textbf{(0.0/77.0/23.0)}\end{tabular} \\
 & \textit{Diversity} & \begin{tabular}[c]{@{}c@{}}3.2/90.4/6.4\\\textbf{(2.8/75.8/21.4)}\end{tabular} & \begin{tabular}[c]{@{}c@{}}0.0/95.2/4.8\\\textbf{(0.0/80.2/19.8)}\end{tabular} & \begin{tabular}[c]{@{}c@{}}1.6/92.0/6.4\\\textbf{(1.3/77.3/21.4)}\end{tabular} & \begin{tabular}[c]{@{}c@{}}1.6/92.8/5.6\\\textbf{(1.4/78.0/20.6)}\end{tabular} \\
 & \textit{Empowerment} & \begin{tabular}[c]{@{}c@{}}11.2/74.4/14.4\\\textbf{(9.3/61.3/29.4)}\end{tabular} & \begin{tabular}[c]{@{}c@{}}3.2/86.4/10.4\\\textbf{(2.7/71.9/25.4)}\end{tabular} & \begin{tabular}[c]{@{}c@{}}4.8/79.2/16.0\\\textbf{(4.0/65.0/31.0)}\end{tabular} & \begin{tabular}[c]{@{}c@{}}1.6/91.2/7.2\\\textbf{(1.4/76.4/22.2)}\end{tabular} \\
 & \textit{Overall} & \begin{tabular}[c]{@{}c@{}}0.8/84.8/14.4\\\textbf{(0.7/69.9/29.4)}\end{tabular} & \begin{tabular}[c]{@{}c@{}}0.0/90.4/9.6\\\textbf{(0.0/75.4/24.6)}\end{tabular} & \begin{tabular}[c]{@{}c@{}}0.0/90.4/9.6\\\textbf{(0.0/75.4/24.6)}\end{tabular} & \begin{tabular}[c]{@{}c@{}}0.0/94.4/5.6\\\textbf{(0.0/79.4/20.6)}\end{tabular} \\
\bottomrule
\end{tabular}
}
\caption{Pairwise win-rate comparison under two judge models. Each entry is formatted as \textit{Other Method Win / MegaRAG Win / Tie}. The main value denotes the GPT-4.1-mini result, and the value in parentheses denotes the Gemini-3-Flash result.}
\label{tab:appendix_full_winrate_judges}
\end{table*}
\begin{table}[t]
\centering
\small
\setlength{\tabcolsep}{5pt}
\begin{tabular}{lcc}
\toprule
\textbf{Metric} & \textbf{World History} & \textbf{Env.\ Report} \\
\midrule
Comprehensiveness & 9.2 / 72.8 / 18.0 & 6.7 / 84.5 / 8.8 \\
Diversity         & 15.6 / 70.8 / 13.6 & 12.4 / 74.8 / 12.8 \\
Empowerment       & 18.3 / 62.5 / 19.2 & 14.7 / 69.7 / 15.6 \\
Overall           & 8.8 / 71.2 / 20.0 & 5.5 / 84.1 / 10.4 \\
\bottomrule
\end{tabular}
\caption{Pairwise LLM-judge win rates (\%) between Query-driven MMKG and MegaRAG. Each entry is formatted as \textit{Query-driven Win / MegaRAG Win / Tie}.}
\label{tab:appendix_querydriven_results}
\end{table}

\subsection{Analysis of Evaluation Biases}
\label{sec:appendix_eval_bias}

We further analyze two potential sources of bias in our LLM-based evaluation, namely \textit{positional bias} and \textit{single-model bias}.

\noindent\textbf{Positional Bias.}
To mitigate positional bias, we conduct pairwise evaluation twice by swapping the order of the two candidate answers. If the judge selects the first answer in both rounds, we treat the outcome as a tie, since this behavior likely reflects a positional preference rather than a genuine content-based preference.

\noindent\textbf{Single-Model Bias.}
To reduce dependence on a single judge model, we additionally use Gemini-3-Flash alongside GPT-4.1-mini. The results show that Gemini-3-Flash is more conservative, yielding a tie rate that is approximately 15 percentage points higher on average. Nevertheless, the relative performance rankings remain consistent across the two judges. Moreover, the inter-judge agreement between GPT-4.1-mini and Gemini-3-Flash on winning samples reaches a Cohen's kappa of 0.72, suggesting that MegaRAG's advantages are stable across evaluators.

Table~\ref{tab:appendix_full_winrate_judges} reports the pairwise win-rate comparison. Each entry is formatted as \textit{Other Method Win / MegaRAG Win / Tie}, where the main number denotes the GPT-4.1-mini result and the value in parentheses denotes the Gemini-3-Flash result.

\subsection{Ablation Study on Using GPT-4o-mini Only (without MMRAG)}
\label{sec:appendix_answer_gen}
To ensure that GPT-4o-mini has not been exposed to our evaluation datasets during pretraining, and to confirm that it cannot answer questions solely by relying on its internal knowledge, we conduct an additional ablation study. Specifically, we compare MegaRAG against a retrieval-free baseline where answers are generated using GPT-4o-mini without access to any external context or retrieved information. As shown in Table~\ref{tab:no_retrieval}, MegaRAG consistently outperforms the retrieval-free baseline, highlighting the value of combining retrieval with multimodal knowledge to enhance answer quality.

\subsection{Recent Multimodal GraphRAG}
\label{sec:appendix_querydriven_comparison}

We additionally compare MegaRAG with a recent multimodal GraphRAG method, \textit{Query-driven MMKG}~\citep{bu2025query}. Query-driven MMKG constructs query-specific local multimodal knowledge graphs dynamically at inference time, whereas MegaRAG builds a document-level multimodal knowledge graph offline and further refines it to capture cross-page and cross-modal dependencies. As a result, the two methods are designed for different operating regimes: Query-driven is tailored to short-form, query-centered reasoning, while MegaRAG is designed for long-form document understanding and global reasoning.

We attempted to reproduce Query-driven MMKG based on the published paper and released code. However, some implementation details, particularly the prompts used for entity-relation extraction, were not fully specified. Therefore, the following comparison should be interpreted as a best-effort reproduction rather than an exact replication.

Table~\ref{tab:appendix_querydriven_results} reports the pairwise LLM-judge win rates between Query-driven and MegaRAG on two representative long-document benchmarks. Each entry is formatted as \textit{Query-driven Win / MegaRAG Win / Tie}. MegaRAG consistently outperforms Query-driven across all evaluation dimensions on both datasets. We attribute this gap mainly to the difference in design focus: MegaRAG explicitly models document-level structure and performs global refinement, which is particularly beneficial for long-form multimodal documents requiring cross-page evidence aggregation and multi-hop reasoning.

\clearpage

\subsection{Case Studies}
\label{sec:appendix_case_studies}

We present two case studies demonstrating the benefits of our MMKG refinement stage in improving knowledge extraction from visually rich documents. These examples show how refinement enhances multimodal grounding and enables the recovery of global, cross-page relations.

\noindent\textbf{Example of enhanced multimodal relations.} 

\begin{figure*}
    \centering
    \includegraphics[width=0.6\linewidth]{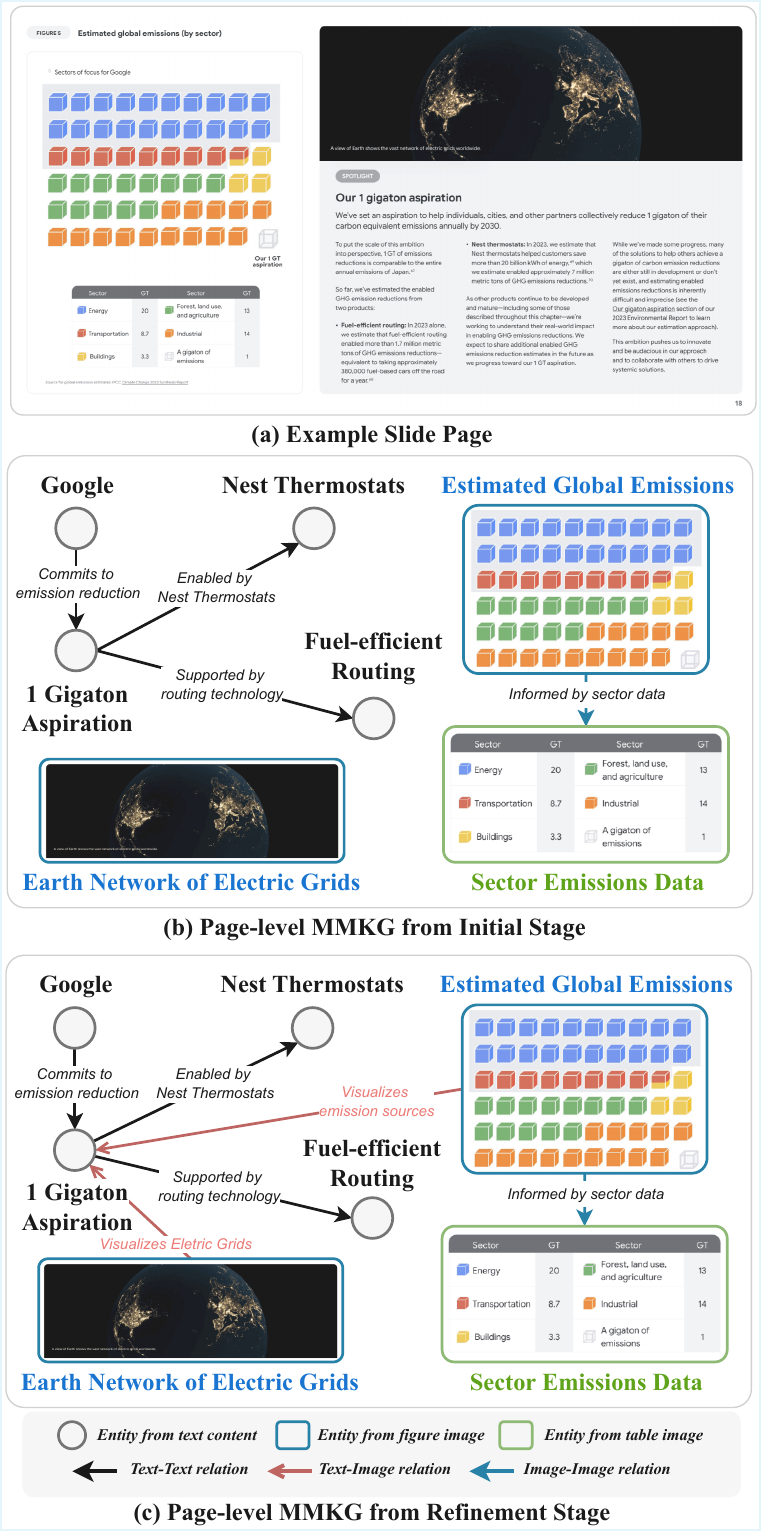}
    \caption{Example of enhanced multimodal relations. (a) A slide page from an environmental report. (b) Page-level MMKG generated in the initial stage. (c) Page-level MMKG from the refinement stage.}
    \label{fig:MMKG_init_and_refine}
\end{figure*}

In the initial MMKG stage shown in Figure~\ref{fig:MMKG_init_and_refine}, 
entities such as \textit{Estimated Global Emissions} and \textit{Earth Network of Electric Grids} are extracted from figure images, but their connections to textual entities are missing. After refinement, these visual entities are correctly linked to the \textit{1 Gigaton Aspiration}.

\noindent\textbf{Example of enhanced cross-page relations.}

\begin{figure*}
    \centering
    \includegraphics[width=0.6\linewidth]{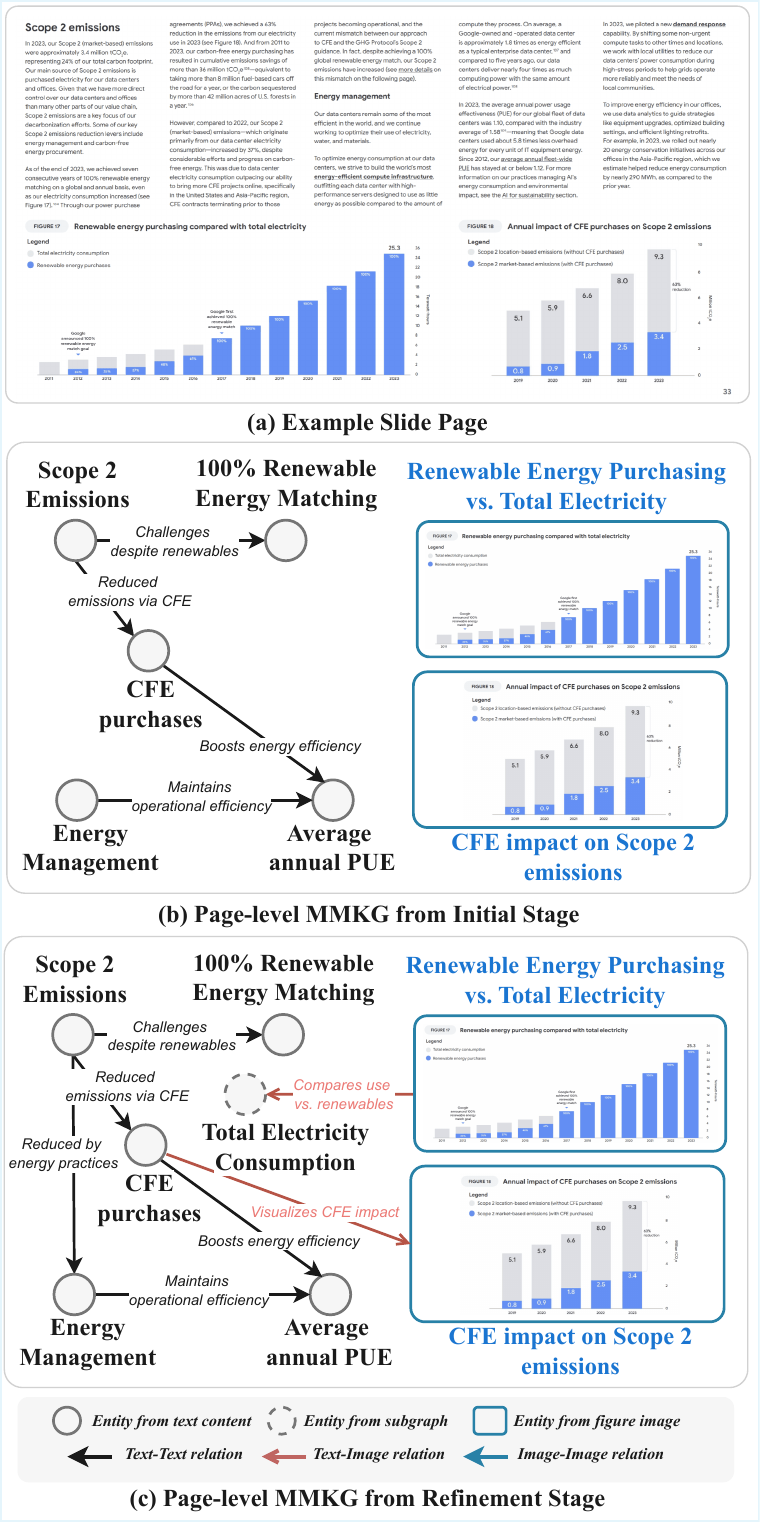}
    \caption{Example of enhanced cross-page relations. (a) A slide page from an environmental report. (b) Page-level MMKG generated in the initial stage. (c) Page-level MMKG from the refinement stage.}
    \label{fig:MMKG_init_and_refine_case2}
\end{figure*}

We deomnstrate that cross-page relations can be recovered after the refinement stage in the example shown in Figure~\ref{fig:MMKG_init_and_refine_case2}. By leveraging the provided MMKG subgraph, our method successfully links the visual entity \textit{Renewable Energy Purchasing vs. Total Electricity}" to the cross-page entity \textit{Total Electricity Consumption}.

\noindent\textbf{Comparative Analysis.}

Further examples are provided in Tables \ref{tab:case_study_q1_graphrag}, \ref{tab:case_study_q1_lightrag}, \ref{tab:case_study_q2_graphrag}, \ref{tab:case_study_q2_lightrag} to compare our MegaRAG with GraphRAG and LightRAG. As shown in the respective LLM judgement, our approach consistently outperforms the baselines across four evaluation metrics: comprehensiveness, diversity, empowerment, and overall.

\newpage

\newpage

\newpage
\renewcommand{\arraystretch}{1.2}
\setlength{\tabcolsep}{8pt}

\begin{table*}[t!]
\centering

\begin{tabular}{p{0.95\textwidth}} 
\toprule
\textbf{Question 1:} What specific carbon reduction strategies are most effective across companies? \\
\midrule

\rowcolor[gray]{0.92}\textbf{\textit{MegaRAG}} \\
Effective Carbon Reduction Strategies Across Companies: The pursuit of effective carbon reduction strategies is critical for companies aiming to mitigate their environmental impact and achieve sustainability goals. Both knowledge from the knowledge graph and insights derived from document images reveal a comprehensive landscape of strategies being implemented by various organizations, highlighting key themes such as emissions reduction, renewable energy initiatives, supply chain management, and investment in carbon removal technologies.

\par \textbf{Decarbonizing Energy Sources:} Leading companies like Meta, Microsoft, and Google are shifting to renewable energy. For instance, Meta has contracted over 11,700 MW of renewable energy to match 100\% of its electricity use. Google has matched its energy use with 100\% renewables for seven years, collaborating with the First Movers Coalition.

\par \textbf{Emissions Reduction Focus:} Meta targets a 42\% reduction in Scope 1 and 2 emissions by 2031. Apple has cut product energy use by over 70\% since 2008. Google and LinkedIn are also working on operational emissions.

\par \textbf{Supplier Engagement:} Microsoft promotes sustainability in its supply chain, encouraging local sourcing and renewable energy use by vendors.

\par \textbf{Investment in Carbon Removal:} Companies like Google and Meta are investing in direct air capture and forest restoration. Apple and Google focus on high-quality carbon credits. Microsoft leads collaborative efforts like Carbon Call to improve reporting standards.

\par \textbf{Conclusion:} Effective strategies span emissions reduction, renewable energy, supplier engagement, and carbon removal investments.

\\
\rowcolor[gray]{0.92}\textbf{\textit{GraphRAG}} \\
\textit{Effective Carbon Reduction Strategies Across Companies:} Companies are pursuing a range of strategies:

\par \textbf{Renewable Energy:} NVIDIA aims for 100\% renewable electricity by FY2025. Apple is committed to carbon neutrality by 2030.

\par \textbf{Scope 3 Emissions:} Google, Apple, and Meta focus on tracking and reducing value-chain emissions.

\par \textbf{Climate Innovation:} Microsoft's Climate Innovation Fund invested \$761M in 2023 to support new technologies.

\par \textbf{Circular Economy:} Apple incorporates recycled materials into products; Meta operates on 100\% renewable energy since 2020.

\par \textbf{Supplier Programs:} Apple’s Supplier Energy Efficiency Program helps reduce emissions across the supply chain.

\par \textbf{Summary:} Key strategies include renewables, Scope 3 management, innovation funding, circularity, and supplier engagement.

\\
\rowcolor[gray]{0.92}\textbf{\textit{LLM Judgement}} \\
\textbf{Comprehensiveness – Winner: MegaRAG} \\
Answer 1 gives a broader, more detailed view, linking specific strategies, company actions, and goals. Answer 2, while informative, is more concise.

\par \textbf{Diversity – Winner: MegaRAG} \\
Answer 1 presents a wider variety of strategies, including both direct and indirect emissions, and a broader range of carbon removal methods.

\par \textbf{Empowerment – Winner: MegaRAG} \\
Answer 1 empowers readers with clear explanations, real examples, and reasoning behind each strategy. Answer 2 lacks the same level of depth.

\par \textbf{Overall – Winner: MegaRAG} \\
Answer 1 is the most comprehensive, diverse, and empowering of the two answers.

\\
\bottomrule
\end{tabular}
\caption{Case (1) Study: Comparison between MegaRAG and GraphRAG.}
\label{tab:case_study_q1_graphrag}
\end{table*}

\renewcommand{\arraystretch}{1.2}
\setlength{\tabcolsep}{8pt}

\begin{table*}[t!]
\centering

\begin{tabular}{p{0.95\textwidth}} 
\toprule
\textbf{Question 1:} What specific carbon reduction strategies are most effective across companies? \\
\midrule

\rowcolor[gray]{0.92}\textbf{\textit{MegaRAG}} \\

Due to space constraints, the detailed responses are presented in Table~\ref{tab:case_study_q1_graphrag} (MegaRAG).

\\
\rowcolor[gray]{0.92}\textbf{\textit{LightRAG}} \\
\textit{Effective Carbon Reduction Strategies Across Companies:} Many companies, particularly in the technology sector, are implementing various carbon reduction strategies to achieve sustainability and mitigate climate change. Here are some of the most effective strategies identified across major organizations like Apple, Google, and Meta:

\par \textbf{Commitment to Renewable Energy:} Companies like Apple have achieved 100\% renewable electricity for operations and are extending this to their supply chains. Meta and Google also match their energy use with renewables to reduce operational emissions.

\par \textbf{Carbon Removals and Offsetting:} Apple and Meta invest in high-quality carbon removal projects to handle residual emissions and support their net-zero goals.

\par \textbf{Efficiency Improvements:} Microsoft improves energy efficiency in data centers. Apple implements operational efficiency practices to lower emissions.

\par \textbf{Supply Chain Engagement:} Apple’s programs help suppliers adopt low-carbon technologies. Meta's Net Zero Supplier Engagement Program supports suppliers in meeting emission reduction targets.

\par \textbf{Innovative Technologies and Practices:} Companies are adopting low-carbon technologies and nature-based solutions to reduce emissions in production and processes.

\par \textbf{Transparency and Reporting:} Apple and Meta publicly report their emissions and progress, increasing accountability and stakeholder trust.

\par \textbf{Conclusion:} These strategies show a unified effort among tech companies to reduce emissions through renewable energy, efficiency, supplier collaboration, and innovation.

\\
\rowcolor[gray]{0.92}\textbf{\textit{LLM Judgement}} \\
\textbf{Comprehensiveness – Winner: MegaRAG} \\
Answer 1 delivers a more detailed view of carbon reduction strategies. It covers decarbonization, emissions targets, supplier engagement, carbon removal, and collaborative efforts, all backed by specific examples and data. Answer 2 provides a solid overview but lacks the same level of depth and specificity.

\par \textbf{Diversity – Winner: MegaRAG} \\
Answer 1 explores a wider range of approaches, from operational reductions to nature- and tech-based carbon removal, and cross-company collaborations. Answer 2 touches on similar themes but offers fewer perspectives and less nuance.

\par \textbf{Empowerment – Winner: MegaRAG} \\
By including concrete commitments, rationales, and cross-sector initiatives, Answer 1 helps readers better understand and assess real-world strategies. Answer 2 is clear but more general, offering less actionable insight.

\par \textbf{Overall – Winner: MegaRAG} \\
Answer 1 stands out for its depth, breadth, and practical value. It consistently outperforms Answer 2 across all criteria, making it a more comprehensive and informative response for understanding corporate carbon reduction efforts.

\\
\bottomrule
\end{tabular}
\caption{Case (1) Study: Comparison between MegaRAG and LightRAG.}
\label{tab:case_study_q1_lightrag}
\end{table*}
\renewcommand{\arraystretch}{1.2}
\setlength{\tabcolsep}{8pt}

\begin{table*}[t!]
\centering

\begin{tabular}{p{0.95\textwidth}} 
\toprule
\textbf{Question 2:} How can generative models be used to augment image classification datasets? \\
\midrule

\rowcolor[gray]{0.92}\textbf{\textit{MegaRAG}} \\
\textbf{Utilizing Generative Models for Augmenting Image Classification Datasets}

Generative models, such as Generative Adversarial Networks (GANs) and Variational Autoencoders (VAEs), enhance image classification datasets by increasing both size and quality.

\par \textbf{Data Generation:} These models create synthetic images that resemble real data. GANs can alter image attributes without extra labeling, while VAEs sample new images from the original data distribution—crucial for small datasets.

\par \textbf{Enhancing Feature Learning:} Variations created by generative models help classifiers recognize subtle differences and learn more discriminative features, improving accuracy under different conditions like lighting or orientation.

\par \textbf{Insight Extraction:} Generative models reveal hidden data structures in high-dimensional spaces—beneficial in fields like medical imaging—helping researchers understand data with limited labeled samples.

\par \textbf{Robust Augmentation:} Advanced augmentation techniques, including super-resolution and colorization, not only create new data but also enhance robustness against overfitting by exposing models to more diverse representations.

\par \textbf{Simulation and Adaptation:} They can simulate diverse environments for applications like robotics and reinforcement learning, allowing broader scenario training for better generalization.

\\
\rowcolor[gray]{0.92}\textbf{\textit{GraphRAG}} \\
Generative models such as GANs and VAEs augment image classification datasets by generating synthetic images, improving dataset size and diversity. This process addresses class imbalance and boosts model robustness.

\par GANs use a generator and discriminator to produce realistic images, which are helpful when the dataset is small or imbalanced. VAEs encode input data into a latent space and sample from it to create new images, increasing dataset variation.

\par Generative models can also generate labeled data, reducing the need for expensive manual annotation. Basic augmentation like color shifts, noise addition, and rotations enhances generalization.

\par By integrating generative models early in the data pipeline, models learn from both real and synthetic images, leading to higher accuracy and reduced overfitting.

\\
\rowcolor[gray]{0.92}\textbf{\textit{LLM Judgement}} \\
\textbf{Comprehensiveness – Winner: MegaRAG} \\
Answer 1 offers a more complete explanation, discussing synthetic generation, feature learning, data structure analysis, augmentation robustness, and simulation. Answer 2 focuses mainly on synthetic data and standard augmentation.

\par \textbf{Diversity – Winner: MegaRAG} \\
Answer 1 explores multiple dimensions—from feature learning to simulation—while Answer 2 concentrates on class imbalance and dataset enlargement.

\par \textbf{Empowerment – Winner: MegaRAG} \\
Answer 1 gives readers a clearer picture of how and why generative models are effective, with practical examples and diverse use cases. Answer 2 is helpful but more limited in depth.

\par \textbf{Overall – Winner: MegaRAG} \\
Answer 1 outperforms Answer 2 in all aspects, providing broader insights and more actionable information for leveraging generative models in dataset augmentation.

\\
\bottomrule
\end{tabular}
\caption{Case (2) Study: Comparison between MegaRAG and GraphRAG.}
\label{tab:case_study_q2_graphrag}
\end{table*}
\renewcommand{\arraystretch}{1.2}
\setlength{\tabcolsep}{8pt}

\begin{table*}[t!]
\centering

\begin{tabular}{p{0.95\textwidth}} 
\toprule
\textbf{Question 2:} How can generative models be used to augment image classification datasets? \\
\midrule

\rowcolor[gray]{0.92}\textbf{\textit{MegaRAG}} \\
Due to space constraints, the detailed responses are presented in Table~\ref{tab:case_study_q2_graphrag} (MegaRAG).

\\
\rowcolor[gray]{0.92}\textbf{\textit{LightRAG}} \\
\textbf{Augmenting Image Classification Datasets with Generative Models}

Generative models enhance image classification datasets by producing synthetic images that resemble original data, which is crucial for improving model robustness and generalization when data is limited.

\par \textbf{Data Generation:} GANs and VAEs can learn the data distribution and sample new images, expanding dataset size and diversity.

\par \textbf{Creating Diverse Samples:} Through transformations like rotation, scaling, color shifts, and noise, these models generate varied images, helping prevent overfitting.

\par \textbf{Enhancing Underrepresented Classes:} Generative models can target low-sample classes to balance the dataset.

\par \textbf{Improving Model Stability:} Adding synthetic data introduces broader scenarios, improving training stability and real-world performance.

\par \textbf{Applications:} Augmented datasets aid CNNs (e.g., AlexNet) and fine-tuning in transfer learning tasks for better classification results.

\\
\rowcolor[gray]{0.92}\textbf{\textit{LLM Judgement}} \\
\textbf{Comprehensiveness – Winner: MegaRAG} \\
Answer 1 provides broader and more detailed coverage, including feature learning, high-dimensional insight extraction, advanced augmentation (e.g., super-resolution), and simulation. Answer 2 covers core concepts well but lacks these deeper applications.

\par \textbf{Diversity – Winner: MegaRAG} \\
Answer 1 discusses a wider array of technical and application perspectives—ranging from data generation to domain-specific use. Answer 2 focuses more narrowly on basic augmentation and dataset balance.

\par \textbf{Empowerment – Winner: MegaRAG} \\
Answer 1 better equips readers by showing *how and why* generative models enhance data. It includes multiple use cases and explains strategic benefits. Answer 2 is more concise, with fewer actionable insights.

\par \textbf{Overall – Winner: MegaRAG} \\
Answer 1 wins across all criteria. Its comprehensive scope, nuanced techniques, and practical guidance make it more informative and valuable overall.

\\
\bottomrule
\end{tabular}
\caption{Case (2) Study: Comparison between MegaRAG and LightRAG.}
\label{tab:case_study_q2_lightrag}
\end{table*}

\end{document}